\documentclass[10pt,twocolumn,letterpaper]{article}

\usepackage{iccv}
\usepackage{times}
\usepackage{epsfig}
\usepackage{graphicx}
\usepackage{amsmath}
\usepackage{amssymb}
\usepackage{caption}
\usepackage{booktabs}
\usepackage{array}
\usepackage{float}

\usepackage[breaklinks=true,bookmarks=false]{hyperref}

\iccvfinalcopy 

\def\iccvPaperID{9771} 

\ificcvfinal\fi

\begin{document}

\title{ArtFusion: Controllable Arbitrary Style Transfer using \\
Dual Conditional Latent Diffusion Models}

\author{Dar-Yen Chen\\
{\tt\small chendaryen@outlook.com}
}


\twocolumn[{%
\renewcommand\twocolumn[1][]{#1}%
\maketitle
\ificcvfinal\thispagestyle{empty}\fi
\begin{center}
    \centering
    \captionsetup{type=figure}
    \includegraphics[width=1.\linewidth]{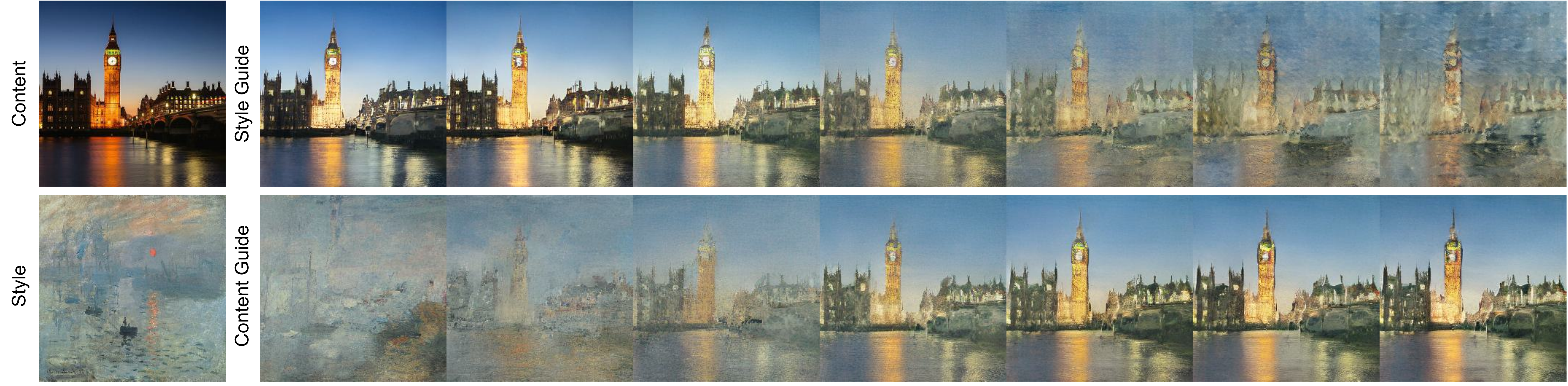}
    \captionof{figure}{Results of our ArtFusion using classifier-free guidance along the style and content conditions. We can adjust the degree of content and style fusion during inference, dynamically ranging from under- to over-stylization. From left to right, the content and style guidance scales are $[0.15, 0.25, 0.5, 1, 2, 3, 4]$ and $[0.15, 0.25, 0.5, 1, 3, 5, 7]$, respectively.}
    \label{fig:one_dim}
\end{center}
}]

\begin{abstract}
Arbitrary Style Transfer (AST) aims to transform images by adopting the style from any selected artwork.
Nonetheless, the need to accommodate diverse and subjective user preferences poses a significant challenge.
While some users wish to preserve distinct content structures, others might favor a more pronounced stylization.
Despite advances in feed-forward AST methods, their limited customizability hinders their practical application.
We propose a new approach, ArtFusion, which provides a flexible balance between content and style.
In contrast to traditional methods reliant on biased similarity losses, ArtFusion utilizes our innovative Dual Conditional Latent Diffusion Probabilistic Models (Dual-cLDM). This approach mitigates repetitive patterns and enhances subtle artistic aspects like brush strokes and genre-specific features.
Despite the promising results of conditional diffusion probabilistic models (cDM) in various generative tasks, their introduction to style transfer is challenging due to the requirement for paired training data.
ArtFusion successfully navigates this issue, offering more practical and controllable stylization.
A key element of our approach involves using a single image for both content and style during model training, all the while maintaining effective stylization during inference.
ArtFusion outperforms existing approaches on outstanding controllability and faithful presentation of artistic details, providing evidence of its superior style transfer capabilities.
Furthermore, the Dual-cLDM utilized in ArtFusion carries the potential for a variety of complex multi-condition generative tasks, thus greatly broadening the impact of our research.
\end{abstract}

\section{Introduction}
The objective of style transfer is to synthesise an image $I_{cs}$ that aptly integrates the content from image $I_c$ with the unique stylistic patterns of a given artistic work, $I_s$.
Seminal work by Gatys \etal \cite{Gatys_2016_CVPR} introduced an optimization-based approach. It iteratively enhances the similarity of content and style features using a pretrained deep neural network.
Despite the influence \cite{ijcai2017p310, https://doi.org/10.48550/arxiv.1701.08893, wang2021rethinking} of this method, it has certain inherent limitations, most notably its time-consuming nature.
This shortcoming sparked a shift towards research into feed-forward networks for direct, rapid stylized $I_{cs}$ generation, initiated by Johnson \etal \cite{10.1007/978-3-319-46475-6_43}.

While style transfer models have progressed from transferring a singular style \cite{10.1007/978-3-319-46475-6_43, 10.1007/978-3-319-46487-9_43, 10.5555/3045390.3045533} or a limited number of styles \cite{Chen_2017_CVPR, dumoulin2017a, 10.1145/3450525, Li_2017_CVPR, 10.1007/978-3-030-11018-5_32, Chen_2021_CVPR, Shen_2018_CVPR} to arbitrary styles \cite{adain, DBLP:journals/corr/abs-2006-09029, Kotovenko_2019_CVPR, wang2020collaborative, sheng2018avatar, WCT-NIPS-2017, wang2020diversified, lin2021drafting, Jing_Liu_Ding_Wang_Ding_Song_Wen_2020, DBLP:journals/corr/GhiasiLKDS17, Yao_2019_CVPR, li2018learning, gu2018arbitrary}, contemporary AST models still grapple with major issues.
Challenges include a lack of adjustable results tailored to user subjective demands, leading to undesirably rigid results, under-stylization, or over-stylization \cite{style_aware_norm} that often disappoint the users.
Furthermore, these models frequently suffer from repetitive artifacts and a poignant loss of artistic details owing to bias in style similarities \cite{adaattn, Chen_2021_CVPR, cast, deng2021stytr2, artflow, iest}.

Diffusion probabilistic models (DM) \cite{ddpm} are growing in popularity in the field of computer vision (CV), celebrated for their high-quality, diverse image generation.
With the incorporation of various inference guidance techniques \cite{diffusion_beat_gan, classifier_free, glide}, conditional DMs (cDMs) can offer flexible control over output results, suggesting a promising pathway for addressing AST's challenges.
Nevertheless, direct training of cDMs for style transfer encounters a roadblock: the necessity for paired data in maximum likelihood learning, a condition unsatisfied in many complex multi-condition generative tasks, including style transfer.
While disentangled inference guidance \cite{kwon2023diffusionbased} and optimization-based algorithms \cite{imagic} attempt to solve this, they require heavy computation and careful hyperparameter tuning.
Hence, we raise the question: Can a cDM be trained effectively for AST?

We present ArtFusion, the first diffusion-based AST model, built upon the latent diffusion model (LDM) \cite{ldm}.
ArtFusion introduces the dual conditional conditional LDM (Dual-cLDM) that treats both content and style as conditions.
During the training phase, our model transforms the style transfer task into a self-reconstruction task while retaining robust stylization capacity during the inference phase.
With likelihood learning, we can avoid biased similarity loss, where the similarity measure does not accurately reflect human perception, and artifacts followed.
This novel approach, coupled with the proposed two-dimensional classifier-free guidance (2D-CFG) during sampling (refer to Fig. \ref{fig:one_dim} and \ref{fig:two_dim}), facilitates balanced control between the content and style, thereby catering to users' subjective preferences effectively.
Furthermore, ArtFusion capitalizes on DM's intrinsic ability to generate diverse and highly coherent stylization, outperforming previous feed-forward approaches in deftly expressing subtle style characteristics and showcasing efficiency over inference-only DM methods.

In summary, we offer the following key contributions:

\begin{enumerate}
    \item ArtFusion, the first diffusion-based feed-forward AST model, provides an effective solution for AST.
    \item Dual-cLDM, which breaks the paired data limitation in cDM training, promises to catalyze advancements in other multi-condition generative tasks.
    \item 2D-CFG offers an adjustable tradeoff between content and style, enhancing the applicability of AST.
    \item Comprehensive experiments demonstrating the effectiveness of our approach, showcasing its ability to faithfully transfer style without bias.
\end{enumerate}

\begin{figure}[t]
\begin{center}
   \includegraphics[width=1.\linewidth]{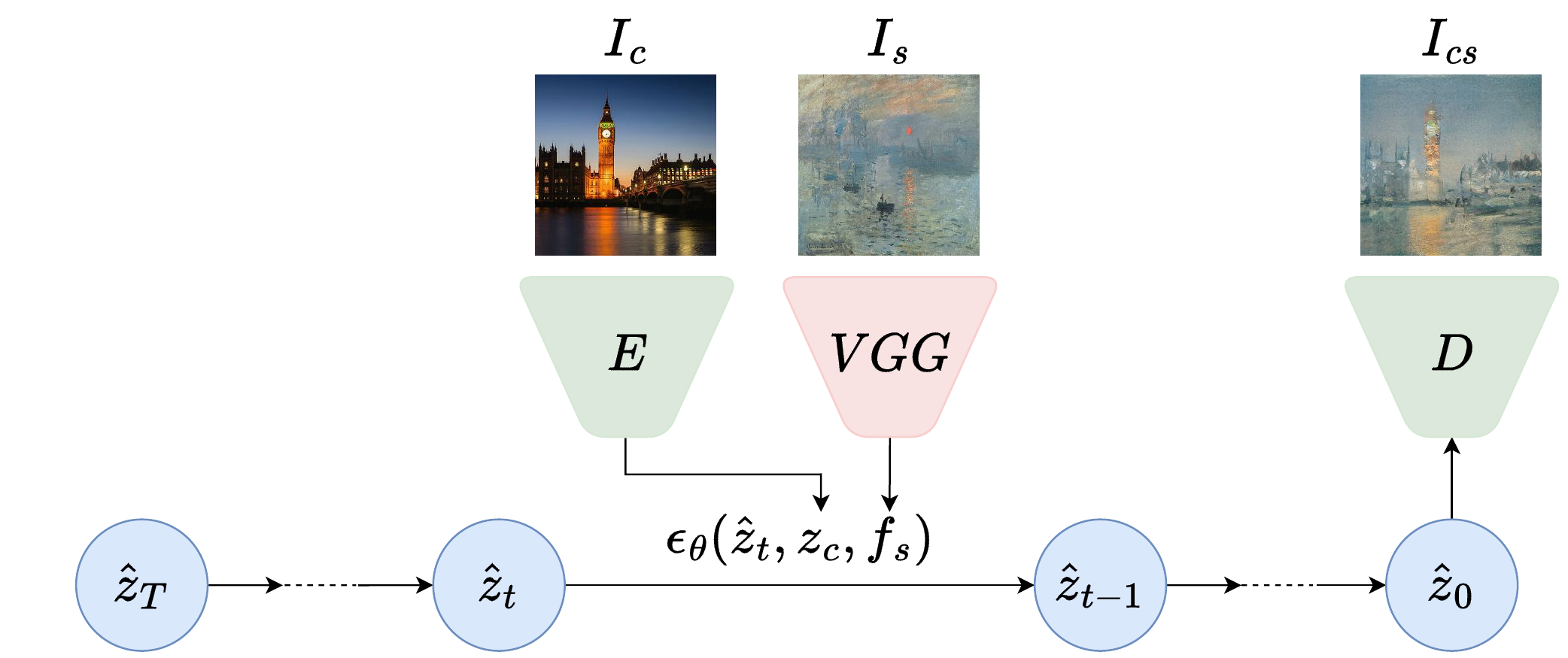}
\end{center}
   \caption{The inference framework of the Dual-cLDM for style transfer.  Initiated from isotropic Gaussian-distributed noise $\hat{z}_T$, the dual conditional backbone progressively denoises using both content and style as conditions. Post-denoising, the $\hat{z}_0$ is decoded using the first-stage decoder.}
\label{fig:denoising}
\end{figure}

\begin{figure*}[t]
\begin{center}
   \includegraphics[width=.995\linewidth]{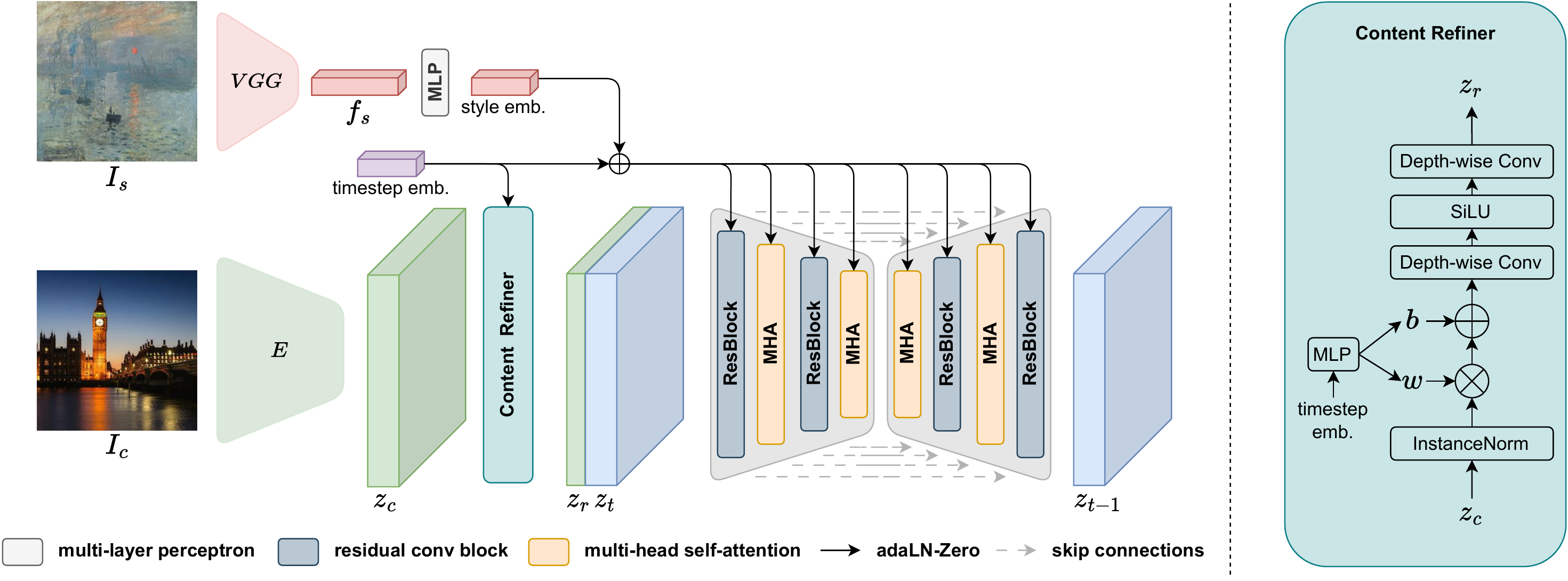}
\end{center}
   \caption{Left: Dual-cLDM Architecture. Pretrained VGG extracts style features $f_s$, while content features $z_c$ are encoded using the first-stage VAE encoder. The content refiner processes $z_c$ into $z_r$, refining content from inherent style. The refined $z_r$ is then concatenated with the noisy latent $z_t$. Style features $f_s$, along with timestep embeddings, are injected via adaptive normalization. Right: The architecture of the Content Refiner. This design aims to reduce the depth dimension of $z_c$.}
\label{fig:stylefusion}
\end{figure*}

\section{Limitations and Biases of Style Similarity}
While pretrained VGG \cite{vgg} has traditionally played a pivotal role in style transfer, it brings with its drawbacks and biases.
Having been trained on natural images for classification tasks, VGG's repurposing for style extraction from artistic images encounters certain obstacles. Specifically, its capabilities in capturing certain aspects of style, such as color hue and geometric patterns, does not effectively extend to more abstract elements crucial to art, such as brush strokes, textures of various painting mediums (\eg oil-painting, watercolor, sketch), or genre nuances.
Such essential artistic features might be infrequent in natural images, or may even be consciously disregarded by classification-oriented models.
As a result, although prior works might show strength with geometric or vivid styles, they falter with more abstract ones, leading to a loss of intricate art details. Consequently, the output might deviate from the intended stylistic vision, restricting its adaptability to varied artistic demands.

Beyond the inherent issues of classification-pretrained VGG models, another limitation arises from the widespread use of second-order statistics style loss in style transfer.
While beneficial for matching feature statistics, this loss inadvertently encourages repetitive artifacts \cite{iest}.
The second-order statistics mean/variance or Gram matrix approach to style representation captures the statistical distribution of features, but neglects their real distribution and spatial arrangements.
This results in style transfer that may overemphasize certain aspects, particularly dominant textures of the style image, to minimise the statistical similarity.
Consequently, statistics style loss often leads to a lack of global coherence, creates annoying artifacts, and misses subtle artistic characteristics \cite{cast}, once again. 
This points to an implicit and flawed aspect of this conventional choice of objective - it does not explicitly define what constitutes style, relying instead on a statistical measure that is assumed, but not assured, to encapsulate style.

These limitations underscore the necessity for novel approaches in style transfer that can more adeptly handle the complexity and subtlety inherent to artistic styles.

\section{Related Work}

\subsection{Feed-forward Arbitrary Style Transfer}
Feed-forward networks, brimming with potential, have been a focal point of research in Arbitrary Style Transfer (AST).
A myriad of researchers \cite{Deng_Tang_Dong_Huang_Ma_Xu_2021, deng:2020:arbitrary, Kotovenko_2019_ICCV, Kotovenko_2019_CVPR, Park2018ArbitraryST, style_aware_norm, deng2021stytr2, cast, iest, adaattn, adain, styleformer, Svoboda_2020_CVPR, sanakoyeu2018styleaware} have made significant contributions to this field.
Typically,  AST models operate with an objective function describing the similarities between content and style representations of output and input images.
Nonetheless, this approach confronts two key challenges: firstly, the bias in content and style representations, and secondly, the lack of flexibility in output control, inevitably limiting the capability to cater to diverse aesthetic preferences.

Several methods have attempted to counter the bias issue \cite{adaattn, Deng_Tang_Dong_Huang_Ma_Xu_2021, deng:2020:arbitrary, Park2018ArbitraryST, styleformer, deng2021stytr2} by exploring self-attention mechanisms in AST.
Deng \etal \cite{deng2021stytr2} notably developed a pure transformer-based architecture to tackle content bias.
Cheng \etal \cite{style_aware_norm} adjusted the style loss to alleviate style bias.
To improve the quality of stylized images, adversarial loss \cite{gan} has been incorporated into AST \cite{https://doi.org/10.48550/arxiv.1701.07875, https://doi.org/10.48550/arxiv.1511.06434, Isola_2017_CVPR, iest, cast}.
Recently, Chen \etal \cite{iest} and Zhang \etal \cite{cast} have used contrastive learning to mitigate bias from pretrained feature extractors and statistics style loss.
Despite the efficacy of these solutions, the issue of style bias persists as a challenge.
Moreover, there has been minimal improvement in the area of output controllability.
To address these issues, we propose a novel approach that employs diffusion probabilistic models with maximum likelihood learning for AST, eliminating the necessity for computing biased similarities and ensuring remarkably versatile and manipulable outputs.

\subsection{Diffusion Probabilistic Model}
In recent years, diffusion probabilistic models (DM) have gained prestige. They have shown the capacity for generating high-quality images.
As a result, more research \cite{ddim, diffusion_beat_gan, d3pm, ddpm_improved, uniform_diffusion, DBLP:journals/corr/abs-2106-15282, kingma2021on, ddpm_distil, ddim_prog_distil, classifier_free, glide} is being invested in this area.
Among them, controlling the progressive inference process is a significant direction \cite{diffusion_beat_gan, classifier_free}. It provides DM with unprecedented controllability.
On the other hand, Rombach \etal \cite{ldm} and Hu \etal \cite{vq_ddm} integrate VQ-GAN \cite{vqgan} with DM.
This integration allows the dimensional reduction of images through first-stage VAEs, making the denoising process less time-consuming.

Conditional DM (cDM) have found widespread application in numerous generative tasks \cite{sr3, https://doi.org/10.48550/arxiv.2207.00050, saharia2022palette, https://doi.org/10.48550/arxiv.2205.07680, Lugmayr_2022_CVPR, glide, ldm, clip_diffusion}.
SR3 was proposed by Saharia \etal \cite{sr3} for super-resolution. Wang \etal \cite{https://doi.org/10.48550/arxiv.2207.00050} achieved success in semantic synthesis. Inpainting was researched by Lugmayr \etal \cite{Lugmayr_2022_CVPR} and Rombach \etal \cite{ldm}. Furthermore, Nichol \etal \cite{glide} and Rombach \etal \cite{ldm} developed stunning text-to-image diffusion models.
Despite these impressive accomplishments, cDMs are confronted with a significant challenge - the need for paired data for training.
This requirement often poses an obstacle for complex generative tasks like AST that require alignment with multiple conditions, \eg content and style.
Some progress has been made by developing post hoc approaches using pretrained DMs.
For instance, Kwon and Ye \cite{kwon2023diffusionbased} proposed a content/style inference guidance.
Also, Kawar \etal \cite{imagic} presented optimization-based methods.
Nevertheless, these methods demand extensive computational inference resources and carefully tuned hyperparameters.
Our work introduces the pioneering learning-based diffusion model for style transfer tasks, designed to generate stylized images directly, hence significantly enhancing the efficiency and effectiveness of the AST.

\begin{figure*}[h]
\begin{center}
   \includegraphics[width=.625\linewidth]{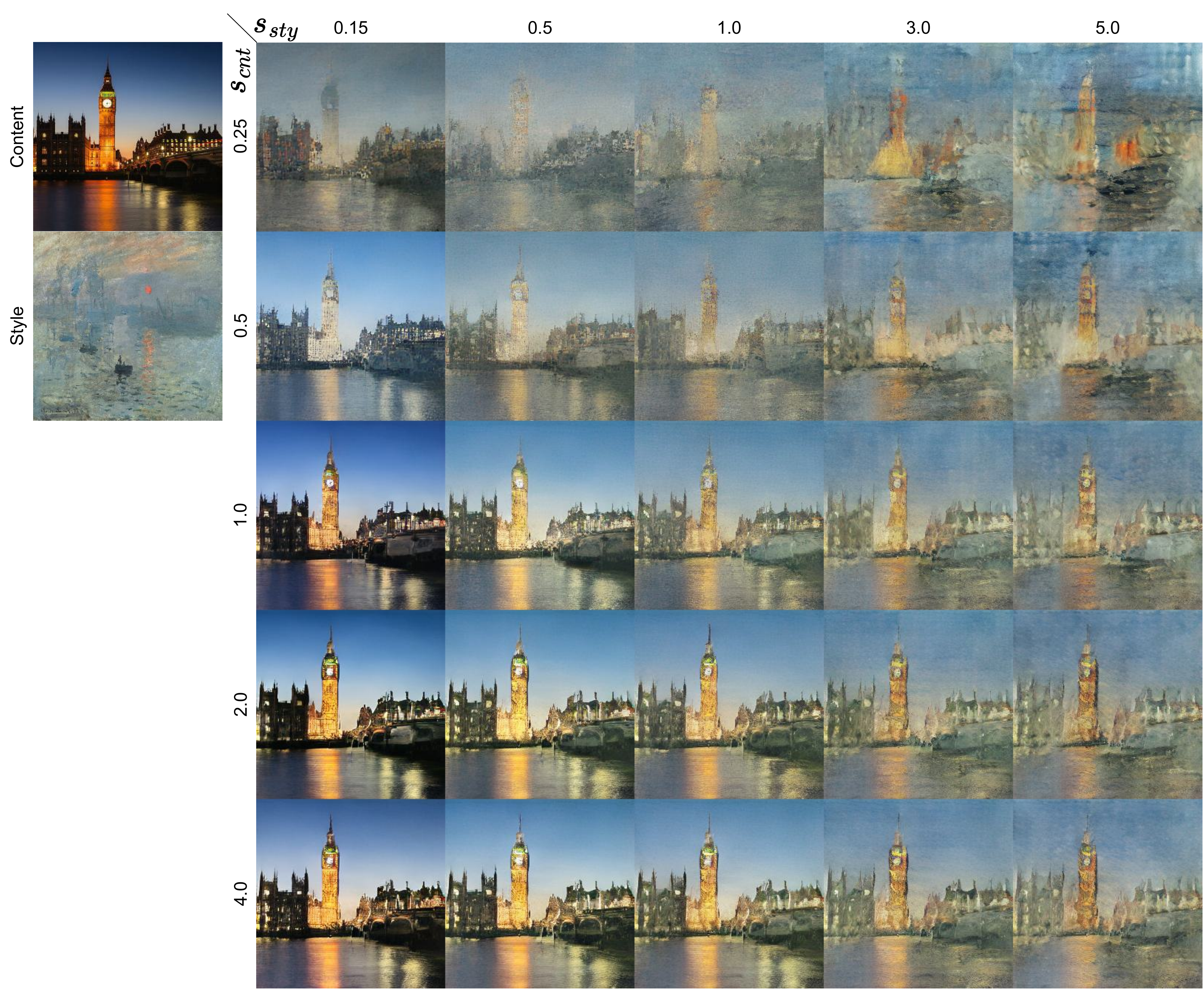}
\end{center}
   \caption{Our 2D-CFG results enable simultaneous content and style manipulation for optimized output, demonstrating flexibility and diversity, thereby offering users with freedom of choice.}
\label{fig:two_dim}
\end{figure*}

\section{Approach}
Our proposed ArtFusion is built on a variant of LDM \cite{ldm}, delivering high-fidelity stylizations that express subtle artistic elements that are often overlooked in previous works.
This is facilitated by a step-by-step denoising process throughout the stylization process (refer to Fig. \ref{fig:denoising}).
Moreover, we empower users with the flexibility to balance between source content and reference style in the outputs, catering to diverse stylization preferences.
Moving forward, this section initially offers an overview of LDM, followed by a detailed explanation of our proposed framework, and its constituent components.
Lastly, we elucidate novel techniques for manipulating the results of stylization.

\noindent\textbf{Preliminaries.}
LDM works with a two-stage framework that combines a VAE and a diffusion backbone.
The key VAE decreases the spatial dimensionality of the image while preserving its semantic essence, resulting in a concise, low-dimensional latent space. The diffusion backbone operates within this latent space, eliminating the need to handle redundant data in the high-dimensional pixel space, thus alleviating the computational burden.
Denote the encoder and decoder of the first-stage VAE as $E$ and $D$ respectively, the image as $I$, and the diffusion backbone as $\epsilon_\theta$.
LDM can be viewed as sequential denoising autoencoders $\epsilon_\theta(z_t,t)$, for $t = 1, ..., T$.
The training objective is to predict the noise at stage $t$ and yield a less noisy version, $z_{t-1}$.
Here, $z_t$ is derived from a diffusion process on $z_0 = E(I)$, this process is modelled as a Markov Chain of length $T$, wherein each step involving a slight Gaussian perturbation of the preceding state.
To keep the notation simple, we will use $\epsilon_\theta(z_t)$ to denote the time-dependent $\epsilon_\theta(z_t,t)$

By applying the reweighted variational lower bound \cite{diffusion_beat_gan}, the objective of LDM become:

\begin{equation}
    \mathcal{L}_{LDM} = \mathbb{E}_{z, \epsilon \sim \mathcal{N}(0,\mathbf{I}), t \sim \mathcal{U}({1, ..., T})}\left[\lVert \epsilon - \epsilon_\theta(z_t) \rVert_2^2\right]
    \label{eq:loss_c_ldm}
\end{equation}

\subsection{Dual Conditional LDM} \label{sec:dual_cldms}
As demonstrated in Fig. \ref{fig:stylefusion}, we establish our approach on the dual conditional LDM (Dual-cLDM) backbone, leveraging a U-Net\cite{unet}-based structure, similar to the one employed in \cite{ldm}. Our training method diverges from conventional style transfer procedures that utilize separate inputs for the content and style images. Instead, a single image serves the dual purpose of providing both the content and the style input, such that $I_{cs} = I_c = I_s$. Consequently, our task shifts from style transfer to self-reconstruction.

\noindent\textbf{First-stage VAE.}
We draw on a pretrained VAE from LDM \cite{ldm}, which has a downsampling factor of 16 and a latent dimension of 16.
Consequently, for an image $I$ with a shape of $3 \times 256 \times 256$, the encoded latent $z = E(I)$ takes on a shape of $16 \times 16 \times 16$.

\noindent\textbf{Conditioning Mechanisms.}
We derive the style feature $f_s$ for the style image $I_s$ by concatenating means and variances from layers within the pre-trained VGG \cite{vgg} network. Propagating $f_s$ through an MLP and subsequently integrating it into the timestep embedding allows us to condition the model using adaLN-Zero \cite{dit}. An intuitive approach for conditioning the content image $I_c$ involves using $z_c := E(I_c)$ as the content feature and combining it with the noisy version $z_t$ via concatenation.
However, an unintended consequence could arise during training. The model might overly rely on $z_c$ and neglect $f_s$, resulting in the compromisation of the stylization ability. This issue stems from the fact that $z_c$ not only contains the content information but also the complete style information.
To circumvent this problem, we introduce a content refiner module that assists the model in refining pure content information from $z_c$.

\noindent\textbf{VGG Style Feature Extractor.}
We utilize the pre-trained VGG-16 \cite{vgg}, which has been trained on ImageNet \cite{imagenet}, to extract features from the style image $I_s$.
The style features are formed by concatenating the means and variances of each feature map in the five style layers \cite{10.1007/978-3-319-46475-6_43} \texttt{relu1\_2}, \texttt{relu2\_2}, \texttt{relu3\_3}, \texttt{relu4\_3} and \texttt{relu5\_3}, which results in a $f_s$ with a length of 2944.

\noindent\textbf{Content Refiner.}
The content refiner, a critical component of our model, serves to refine content and eliminate style from the latent representation $z_c$, producing $z_r$. During training, both content and style are encapsulated within a single image input.
By applying two layers of point-wise convolutions, the content refiner strategically reduces the depth dimension of $z_c$, forcing the elimination of certain information.
Since the model can extract style information from the $f_s$ during training, the content refiner naturally leans towards preserving content while discarding style. Hence, the $z_r$ output is a refined representation, primarily comprising content with lessened style influence. Unless stated otherwise, the content refiner in our approach reduces the original depth dimensions from 16 to 12.

\noindent\textbf{Training Algorithm.}
To harness classifier-free guidance for both content and style, we use shared weights for training the dual conditional and two partial conditional models.
Specifically, $\epsilon_\theta(z_t, z_c, \O_s)$ and $\epsilon_\theta(z_t, \O_c, f_s)$ solely use content or style as condition, respectively. $\O_s$ is the learnable null style, and $\O_c$ is the all zero null content.
Throughout the training, we use probabilities $p_c = 0.1$ and $p_s = 0.5$ for the content-only and style-only models, respectively.

\noindent\textbf{Objective.}
In our training process, $I_{cs} = I_c = I_s$ act as the condition, thus transforming Eqation \ref{eq:loss_c_ldm} into:

\begin{equation}
    \mathcal{L} = \mathbb{E}_{z, \epsilon \sim \mathcal{N}(0,\mathbf{I}), t, I_{cs}}\left[\lVert \epsilon - \epsilon_\theta(z_t, z_c, f_s) \rVert_2^2\right]
\end{equation}

\noindent\textbf{Inference Algorithm.}
The inference denoising process is visually depicted in Fig. \ref{fig:denoising}.
Stylization results are generated by progressively denoising the randomly initialized $\hat{z}_T$ with $\epsilon{\theta}(\hat{z}_t, z_c, f_s)$.
We have provided a detailed explanation of the denoising process in the supplementary materials \ref{sec:details_ddpm}.
Despite being trained for self-reconstruction, our model can still effectively utilize content and style features to achieve remarkable style transfer during inference when fed with different content and style images.

\subsection{Two-Dimensional Classifier-free Guidance}
Earlier feed-forward AST models have developed several manipulation methods, such as style interpolation \cite{adaattn, deng:2020:arbitrary, Park2018ArbitraryST} and spatial control \cite{Park2018ArbitraryST}.
Our proposed model, ArtFusion, not only accommodates these functions but also introduces a more flexible adjustment – the two-dimensional classifier-free guidance (2D-CFG), which is an extension of the classifier-free guidance \cite{classifier_free}.
Using 2D-CFG, users can guide the inference process to lean towards either content or style.
With two scaling factors, $s_{cnt}$ and $s_{sty}$, assigned for content and style respectively, the innovative two-dimensional guidance provides a competitive element in the gradual denoising sampling, driving content and style vie for dominance:

\begin{figure}[t]
\begin{center}
   \includegraphics[width=1.\linewidth]{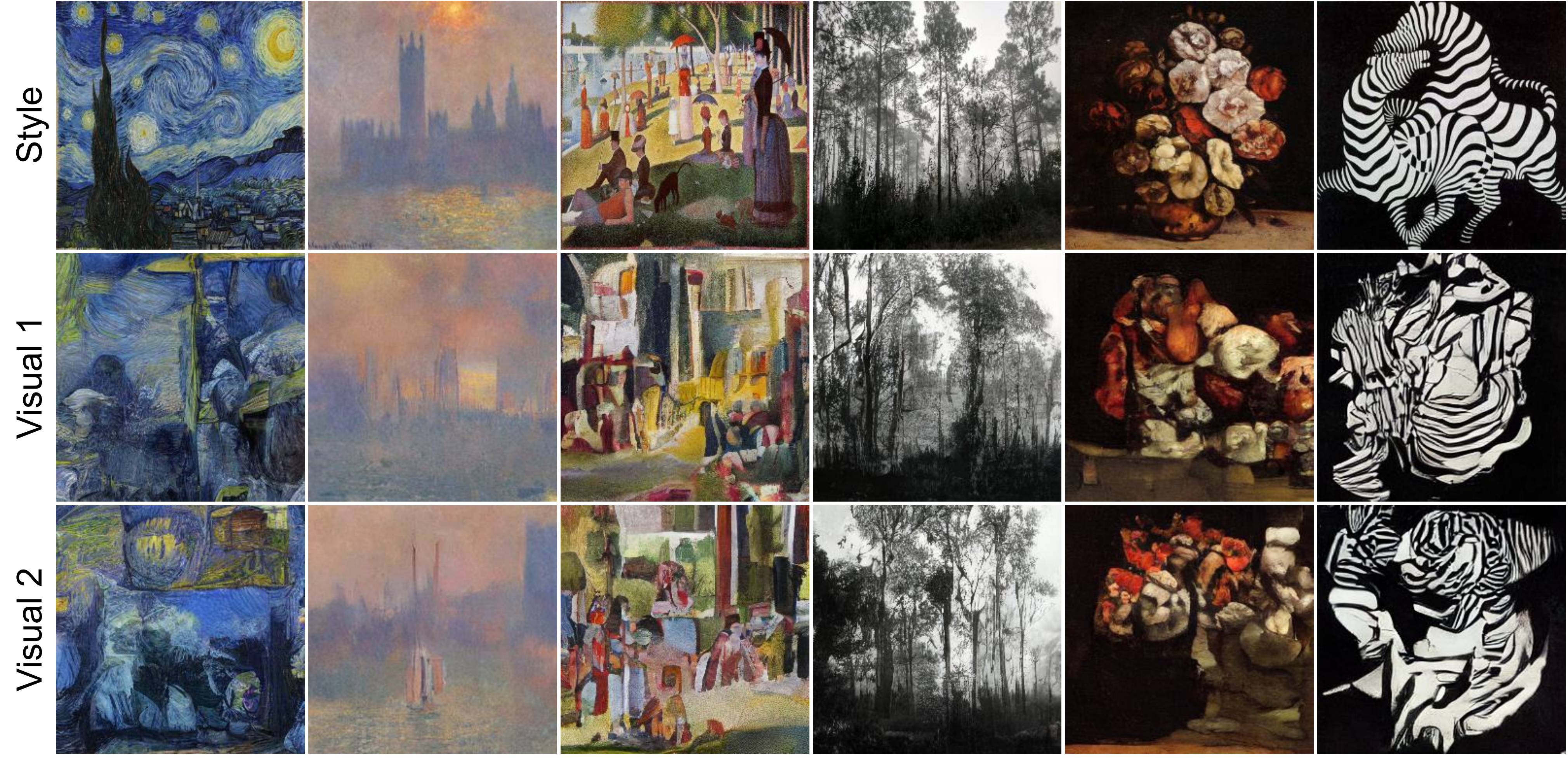}
\end{center}
   \caption{Style visualization from partial $\epsilon_\theta(z_t, \O_c, f_s)$ reveals ArtFusion's faithful expression of style features.}
\label{fig:visulization}
\end{figure}

\begin{figure*}[t]
\begin{center}
   \includegraphics[width=1.\linewidth]{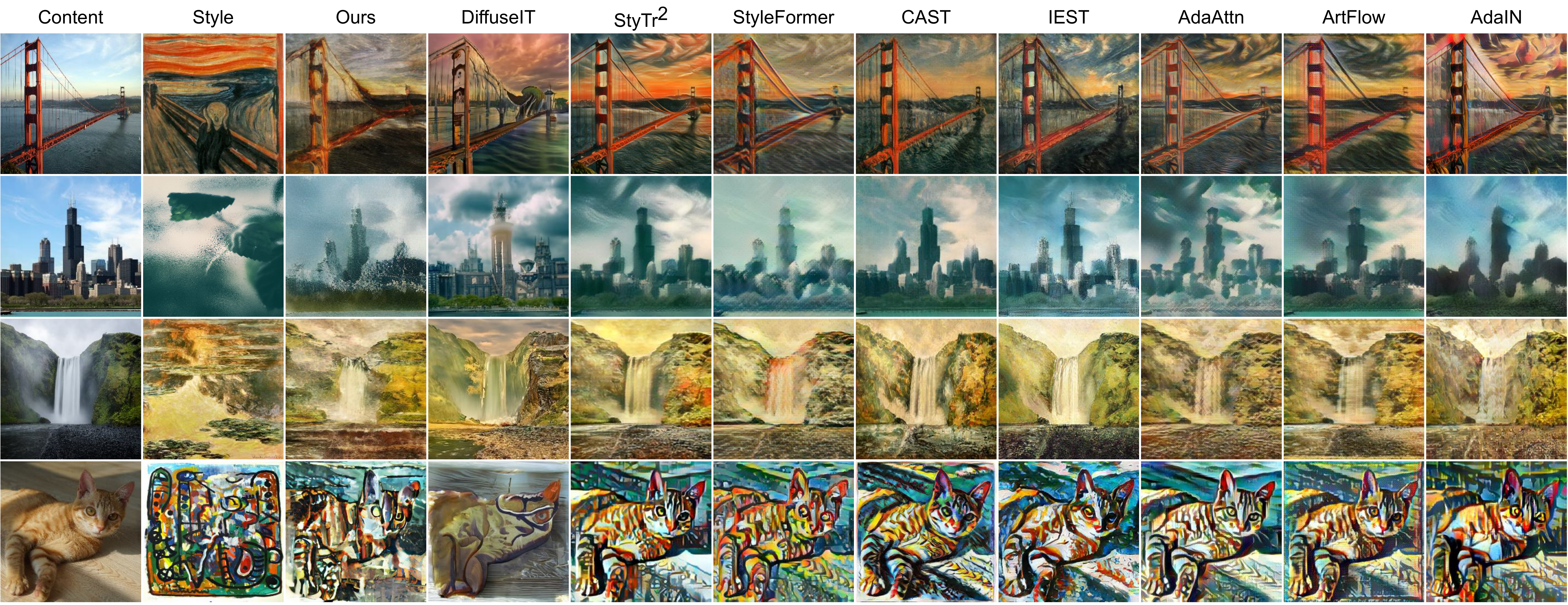}
\end{center}
   \caption{Comparison with SOTA results.}
\label{fig:compares}
\end{figure*}

\begin{equation}
    \tilde{\epsilon}_{\theta, s_{cnt}} = s_{cnt}\epsilon_\theta(\hat{z}_t, z_c, f_s) - (s_{cnt}-1)\epsilon_\theta(\hat{z}_t, \O_c, f_s)
    \label{eq:dual_classifier_free_content}
\end{equation}

\begin{equation}
    \tilde{\epsilon}_{\theta, s_{sty}} = s_{sty}\epsilon_\theta(\hat{z}_t, z_c, f_s) - (s_{sty}-1)\epsilon_\theta(\hat{z}_t, z_c, \O_s)
    \label{eq:dual_classifier_free_style}
\end{equation}

\begin{equation}
    \tilde{\epsilon}_{\theta}(\hat{z}_t, z_c, f_s) = \tilde{\epsilon}_{\theta, s_{cnt}} + \tilde{\epsilon}_{\theta, s_{sty}} - \epsilon_\theta(\hat{z}_t, z_c, f_s)
    \label{eq:classifier_free_dual}
\end{equation}

\section{Experiments}

\subsection{Qualitative Evaluation}
In this section, we evaluate the controllability and fidelity of style reproduction in our proposed method, ArtFusion.
ArtFusion demonstrates controllability by enabling stylization level adjustments.
This aspect is illustrated in Fig. \ref{fig:one_dim}, which presents a spectrum of stylization ranging from vivid content to strong stylization.
Additionally, ArtFusion's two-dimensional classifier-free guidance offers an unprecedented level of nuanced output adjustments.
This capability is showcased in Fig. \ref{fig:two_dim}, where ArtFusion concurrently manipulates content and style, thus easily adapting to various preferences.
Moreover, ArtFusion demonstrates proficiency in integrating style characteristics with the content, thereby yielding striking style transfer results.
Apart from controlling capacities, ArtFusion also manifests talent in style representation.
The model adeptly integrates distinctive style characteristics, such as the blurry edges typical of Impressionist art, with the content to yield compelling results.
Fig. \ref{fig:visulization} serves as proof of this ability, showcasing ArtFusion's faithfulness in style representation.

Our style-conditional model, $\epsilon_\theta(z_t, \O_c, f_s)$, is central to this process.
This model learns to reconstruct $I_s$ independently of $z_c$ content information, establishing a correspondence with the arrangement of style inputs.
For common patterns in the dataset, this link becomes more pronounced.
As exemplified in Fig. \ref{fig:visulization}, the results are logical, especially observable in the depiction of castles in the $2^{nd}$ row, and bottom-up growing trees in the $4^{th}$ row.
By learning the likelihood, our model can grasp the essential traits of various elements, moving closer to comprehending the essence of "real art," a fundamental challenge in style transfer.

\begin{figure}[t]
\begin{center}
   \includegraphics[width=1.\linewidth]{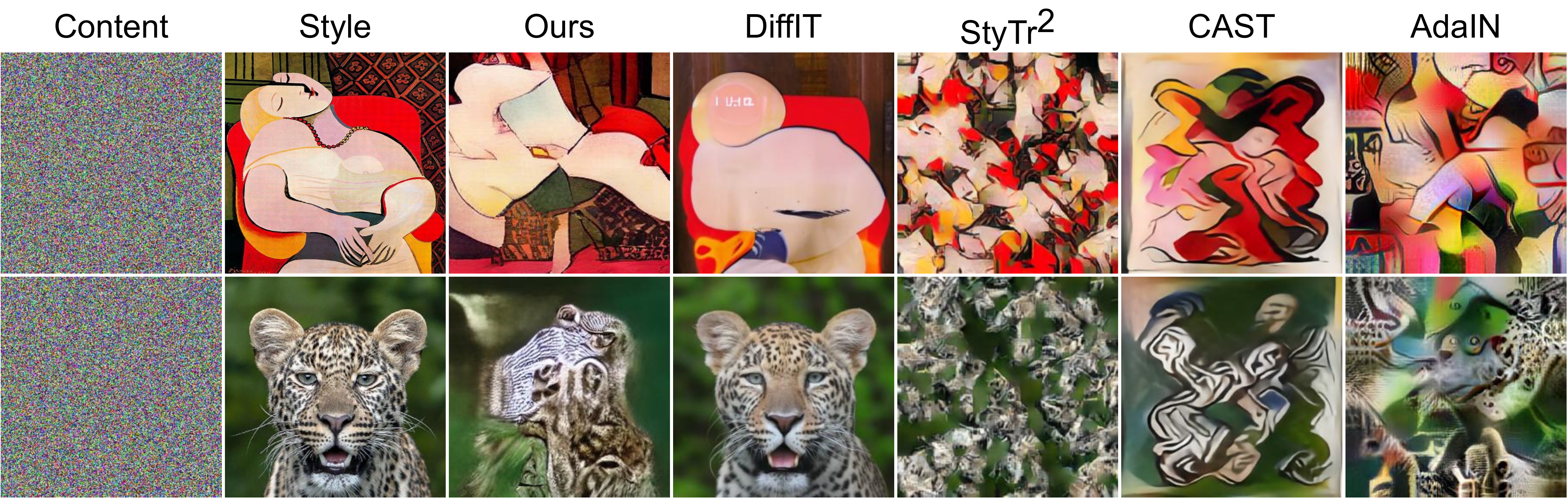}
\end{center}
   \caption{Comparison with style learned by SOTA, using noise content and 20 rounds stylization. ArtFusion avoids repetitive patterns and demonstrates faithful depictions.}
\label{fig:leak}
\end{figure}

\begin{figure*}[t]
\begin{center}
   \includegraphics[width=1.\linewidth]{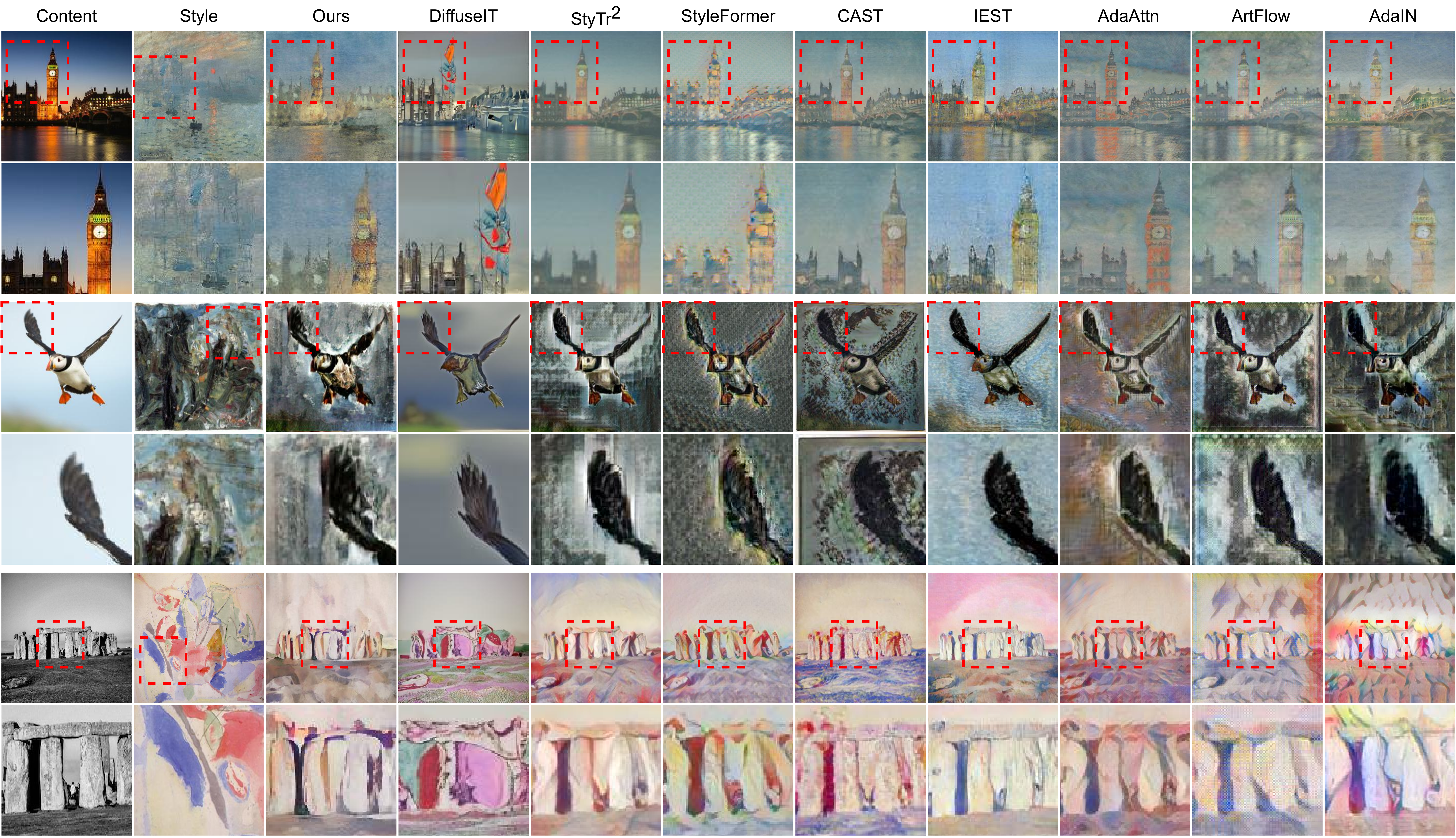}
\end{center}
   \caption{Close-up view of ArtFusion's superior fine style texture representation compared to SOTA models. Each example's second row provides a magnified view.}
\label{fig:compares_detail}
\end{figure*}

\begin{table*}[t]
\begin{center}
    \begin{tabular}{cccccccccc}
        \toprule
         & Our  &   DiffuseIT & StyTr$^{2}$ & StyleFormer & CAST & IEST & AdaAttn & ArtFlow & AdaIN \\
        \midrule
        $\mathcal{L}_{\mu/\sigma}$   & 1.140 &  1.190 & 0.498 & 0.526 & 0.849 & 0.793 & 0.498 & 0.245 & 0.272 \\
        \bottomrule
    \end{tabular}
\end{center}
\caption{VGG style similarity of last example (Stonehenge) in Fig. \ref{fig:compares_detail}. This highlights the inconsistency between visual perception and conventional style similarity.}
\label{tab:bias}
\end{table*}

\subsection{Comparison}
In this section, we compare our method, ArtFusion, with DiffuseIT \cite{kwon2023diffusionbased}, the cutting-edge diffusion-based approach, and seven representative feed-forward AST models: StyTr$^2$ \cite{deng2021stytr2}, Styleformer \cite{styleformer}, CAST \cite{cast}, IEST \cite{iest}, AdaAttn \cite{adaattn}, ArtFlow \cite{artflow}, and AdaIN \cite{adain}.
On the basis of maintaining the content semantics, the comparison is focused on the criteria: alignment with style references.
DiffuseIT  \cite{kwon2023diffusionbased}, although innovative in its use of pretrained DMs and DINO ViT \cite{dino} similarities, tends to struggle with content and style degradation.
This results in inferior outcomes to other models.
The feed-forward AST models \cite{deng2021stytr2, styleformer, cast, iest, adaattn, artflow, adain} often demonstrate a noticeable bias in style representation, leading to a divergence between their generated outputs and original artworks.
The presence of repetitive artifacts and a lack of style texture detail in their generated outputs, as displayed in Fig. \ref{fig:compares_detail}, support this observation.

In contrast, ArtFusion effectively learns the correlation between style conditions and actual artworks, leading to results with superior alignment to style references.
It can capture the core style characteristics that are typically overlooked in prior similarity learning models.
Enlarged details in Fig. \ref{fig:compares_detail} reveal the original-like impression, the texture of oil painting, and similar brush strokes in our results.

\subsection{Comparison on Style}
Fig. \ref{fig:leak} presents a comparative analysis between ArtFusion and other models, with a focus on how each model comprehends and reproduces styles.
We intensify the style and minimise the content impact by applying noise content and 20 consecutive stylization rounds.
DiffuseIT \cite{kwon2023diffusionbased} employs the [CLS] token of DINO ViT for style similarity, which leads to a strong content structure and results that closely resemble style images. This reveals a limitation in the current DINO ViT similarity - an inability to effectively separate content from style, a critical requirement for style transfer.
The pretrained VGG style similarity \cite{deng2021stytr2, adain} prompts models to replicate major patterns from the style reference, leading to the creation of repetitive artifacts and obstructing the capture of subtle style components.
CAST \cite{cast} replaces the statistics similarity with contrastive learning, but still shows a significant style bias and struggles to grasp unique characteristics.
The outcomes tend to be structurally alike, indicating an issue within contrastive learning.

ArtFusion, in contrast, sidesteps these issues. It shuns repetitive patterns and heavy content contexts, resulting in outputs that authentically reflect the style references.
This affirms the capacity for unbiased style learning, accentuating its unique advantage in style transfer.
Attributed to the iterative nature of the denoising process, ArtFusion is able to capture the fine-grained details and essence of style references, resulting in a high-fidelity representation of styles.

\subsection{Analysis of Bias in Style Similarity} \label{sec:analysis}
We examine the alignment of visual perception with quantification results, specifically focusing on the commonly used pretrained VGG mean/variance style loss $\mathcal{L}_{\mu/\sigma}$ \cite{adain}.
This evaluation is represented in Fig. \ref{fig:compares_detail} and Tab. \ref{tab:bias}.
Remarkably, the $\mathcal{L}_{\mu/\sigma}$ metric produces results that do not align with visual quality.
Despite their seemingly superior loss scores, both AdaIN \cite{adain} and ArtFlow\cite{artflow} generate stylized images marked by substantial color distortion and spurious artifacts, deviating visually from the style reference.
In contrast, our model presents the unique brush touch of the style reference, not seen in other models, and stays faithful to other aspects of the style.
However, our model incurs a higher loss, comparable to that of DiffuseIT \cite{kwon2023diffusionbased}, which displays clear distortion.
These findings underline the discrepancy between $\mathcal{L}_{\mu/\sigma}$ and the perceptual quality, serving as a caution against the exclusive reliance on style similarity for achieving optimal style transfer results.

\begin{figure}[t]
\begin{center}
   \includegraphics[width=1.\linewidth]{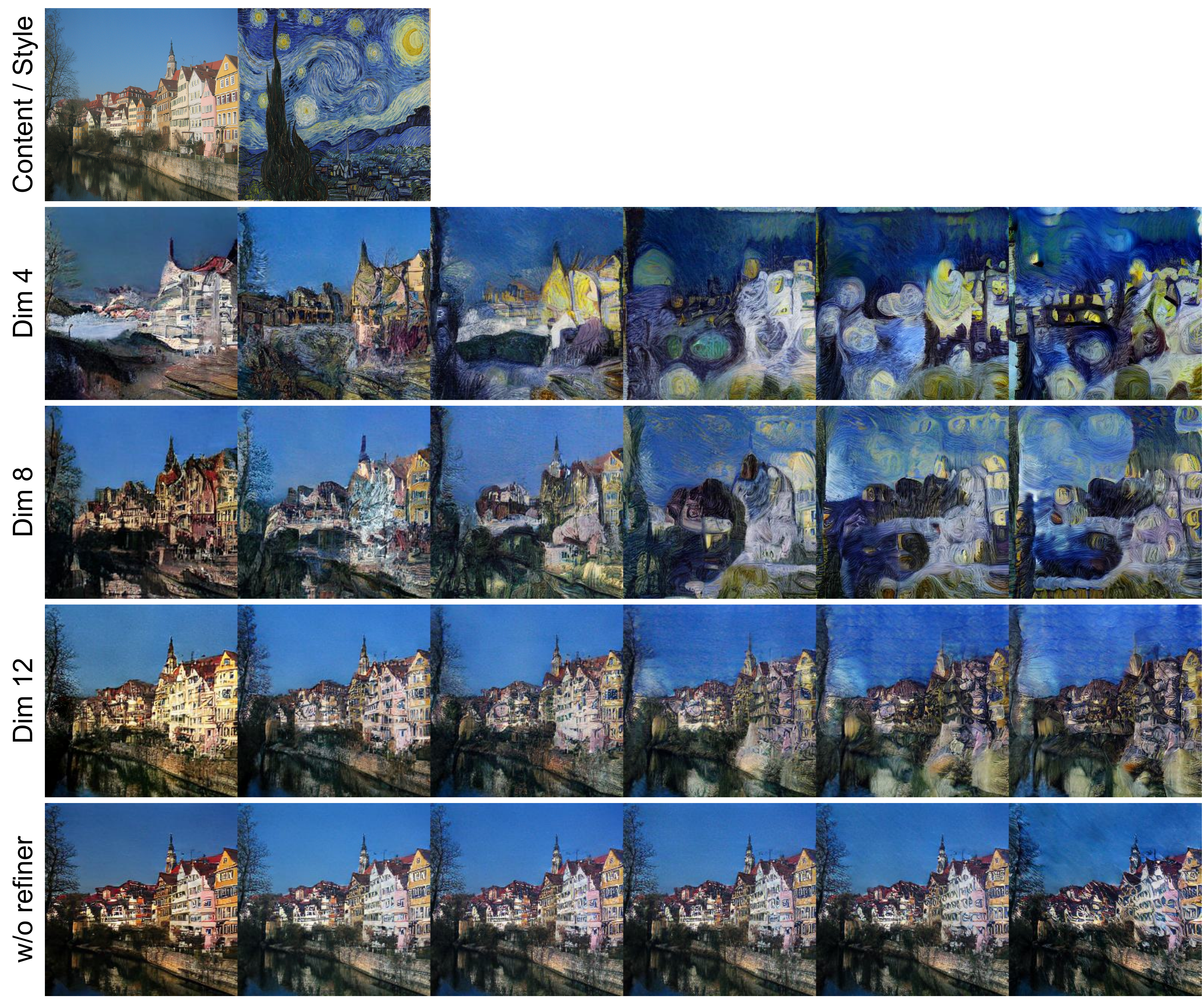}
\end{center}
   \caption{Impact of compression ratio in the content refiner.
Only the content refiner retains content and eliminates style in $z_c$, the model can effectively rely on $f_s$ for style transfer.
The style guidance scale increases from left to right.}
\label{fig:refiner}
\end{figure}

\subsection{Ablation on Content Refiner}
In this section, we investigate the impact of the content refiner on stylization output under varying compression levels of the content feature $z_c$ (Fig. \ref{fig:refiner}).
"Compression" here denotes the reduction of the depth dimension of $z_c$
Analysis indicates that the absence of compression causes $z_c$ to carry an excess of style information, which should ideally be contributed by the style features $f_s$.
As a result, the model over-relies on $z_c$ and is unable to utilize the style information from $f_s$, causing a significant drop in stylization performance during inference (see the $5^{th}$ row in Fig. \ref{fig:refiner}).
On the other hand, over-compression of the content feature leads to inadequate preservation of content semantics, noticeable as an apparent loss of content structure in the output (see the $2^{nd}$ and $3^{rd}$ rows in Fig. \ref{fig:refiner}).
Based on our empirical results, compressing the 16-dimensional $z_c$ down to a 12-dimensional $z_r$ achieves an optimal equilibrium.

\subsection{Interpolation}
\begin{figure}[t]
\begin{center}
   \includegraphics[width=1.\linewidth]{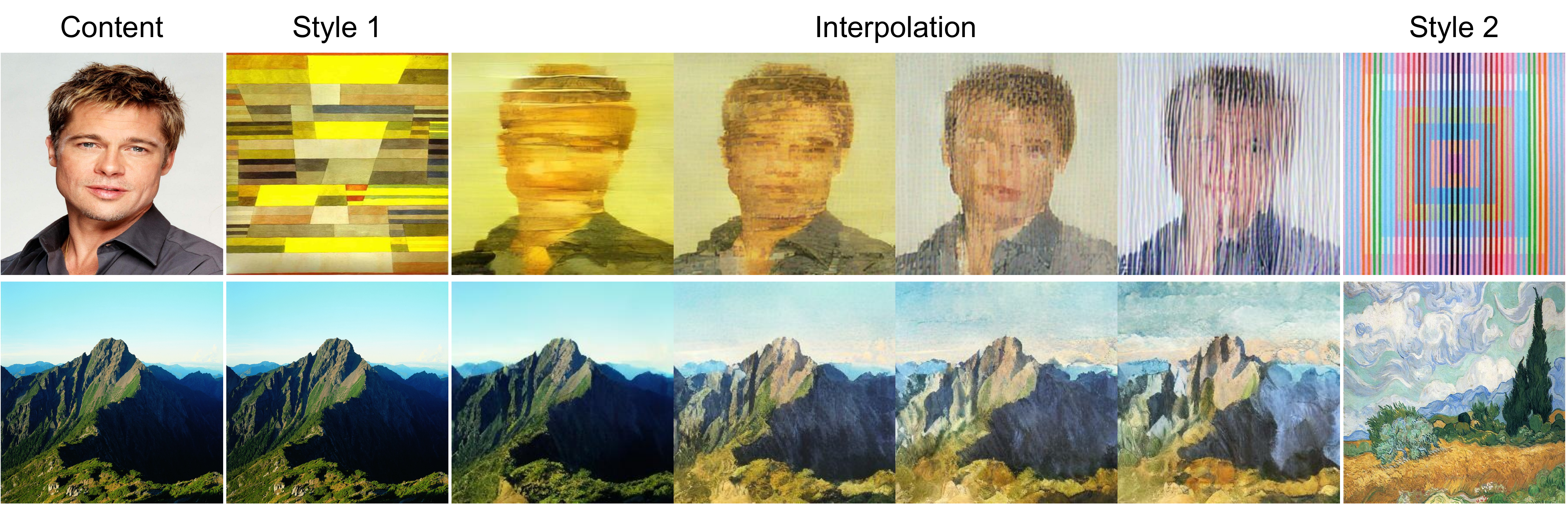}
\end{center}
   \caption{Smooth style interpolation between two styles.}
\label{fig:interpolate}
\end{figure}

Interpolating predicted noise between styles in each intermediate latent space allows for a smooth, gradual shift from one artistic style to another.
Visual examples of the two-dimensional interpolation process are provided in Fig. \ref{fig:interpolate}.
This process enables the seamless blending of artistic features, allowing for the creation of unique, hybrid styles.
Furthermore, when we utilize the content image itself as one of the styles, a one-dimensional content-style tradeoff is introduced.
This spectrum empowers users to finetune the balance between retaining content and adopting a new style, further showcasing the model's outstanding versatility.

\section{Conclusion}
We have presented ArtFusion, a novel, controllable approach to arbitrary style transfer (AST) that leverages dual conditional latent diffusion models (Dual-cLDM).
This framework overcomes the common data limitations associated with diffusion models and avoids biases in feature extractors. With this innovation, ArtFusion effectively expresses and mirrors unique artistic attributes derived from style inputs.
ArtFusion introduces a new level of flexibility to AST through two-dimensional classifier-free guidance (2D-CFG) and noise interpolation.
Significantly, our results demonstrate that ArtFusion effectively prevents common issues found in existing models, such as repetitive patterns, and excels at reproducing nuanced artistic aspects.

The Dual-cLDM employed in ArtFusion harbors potential applications beyond AST, opening the door to other complex generative tasks, and broadening the horizons for diffusion models.
While our model demonstrates a leap forward in the realm of style transfer, a comprehensive understanding of artistic characteristics continues to be a stimulating challenge for future research and exploration.


\onecolumn
\raggedbottom
\renewcommand\thesection{\Alph{section}}
\setcounter{section}{0}


\begin{center}
    {\Large \bf Supplementary Material}
\end{center}

\vspace{0.5cm}

\begin{figure}[h]
\begin{center}
   \includegraphics[width=1.\linewidth]{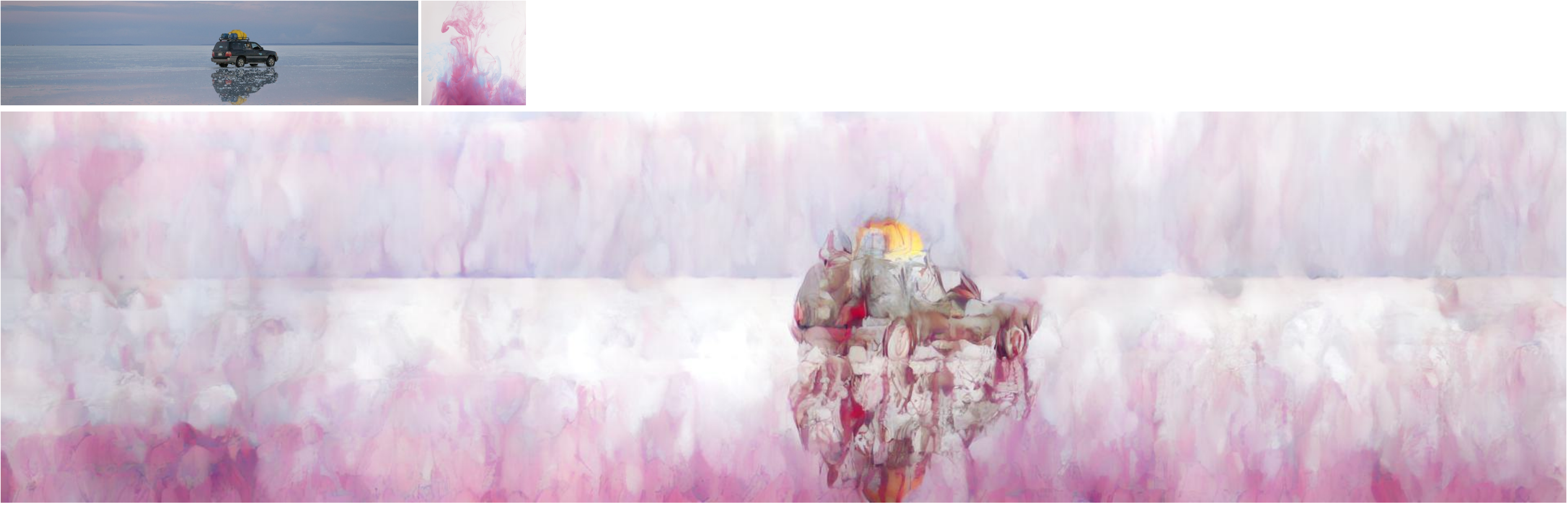}
   
   \vspace{0.5cm}
   
   \includegraphics[width=1.\linewidth]{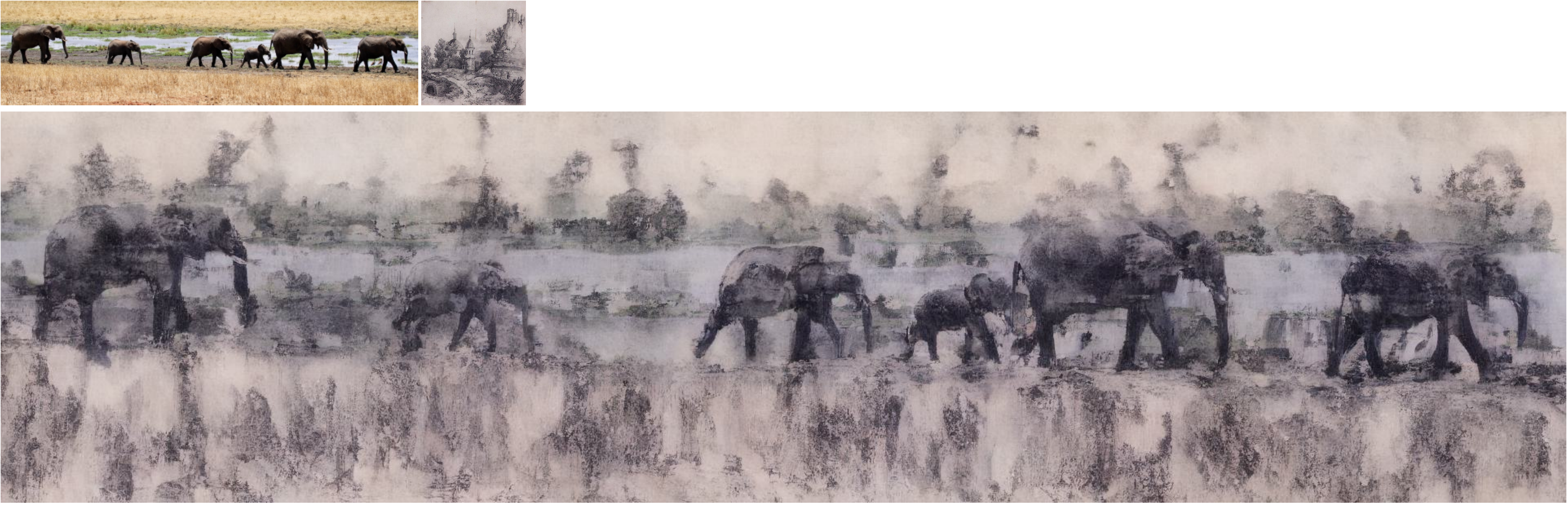}
   
   \vspace{0.5cm}
   
   \includegraphics[width=1.\linewidth]{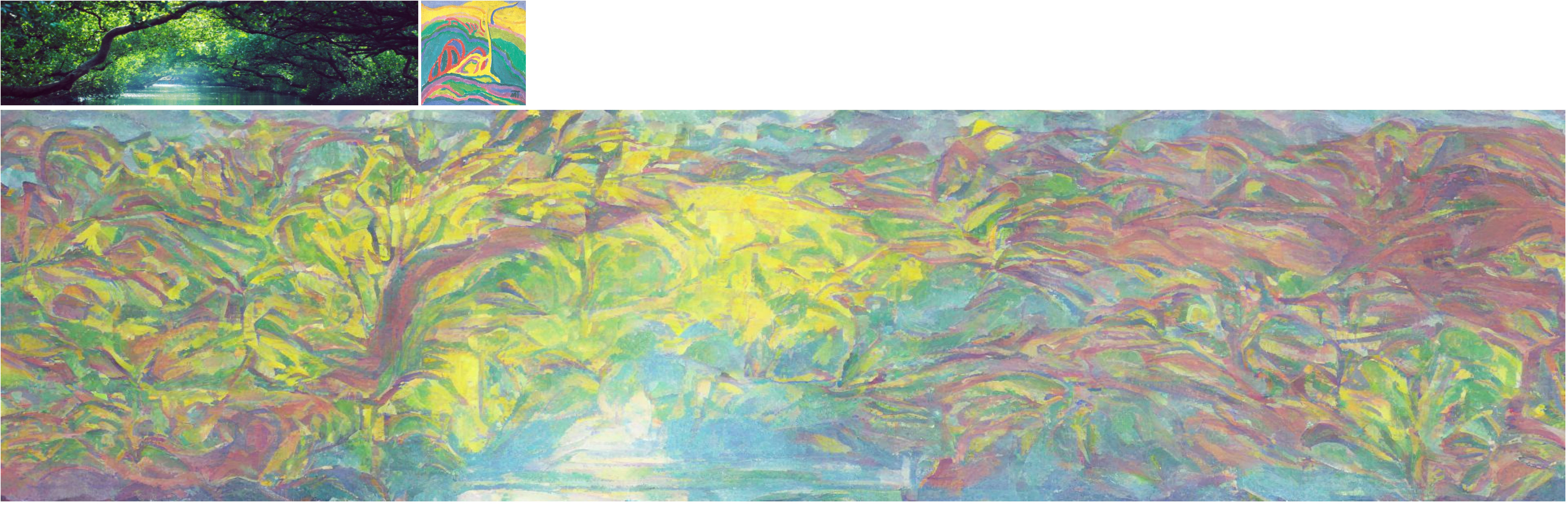}
\end{center}
   \caption{Samples with size $1920 \times 480$. 2D-CFG scales $s_{cnt}/s_{sty}$ from top to bottom: $0.4/3.0$, $0.55/2.0$ and $0.25/2.0$.}
\label{fig:high_resolution}
\end{figure}

\begin{figure}[H]
\begin{center}
   \includegraphics[width=1.\linewidth]{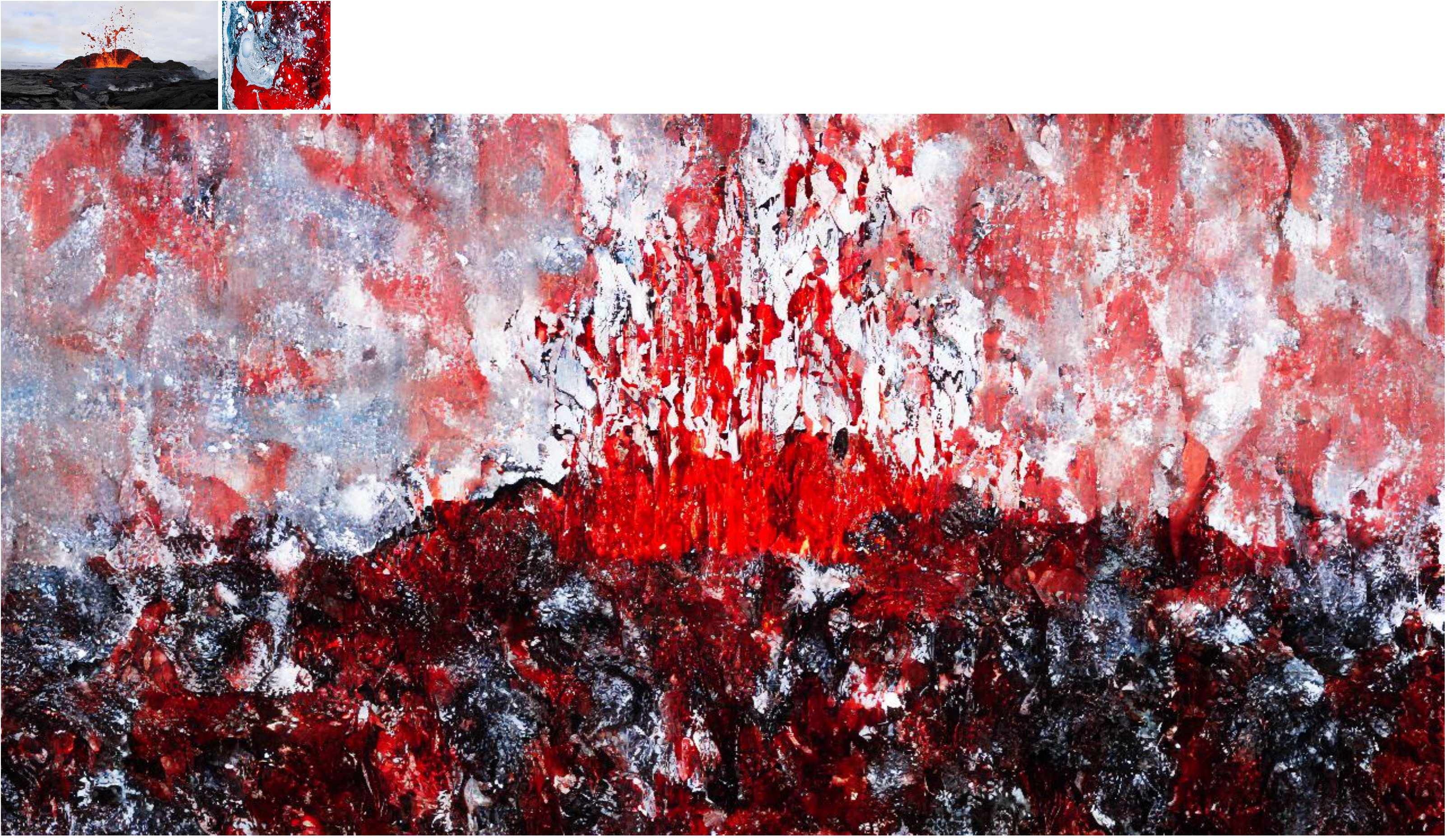}
   
   \vspace{0.5cm}
   
   \includegraphics[width=1.\linewidth]{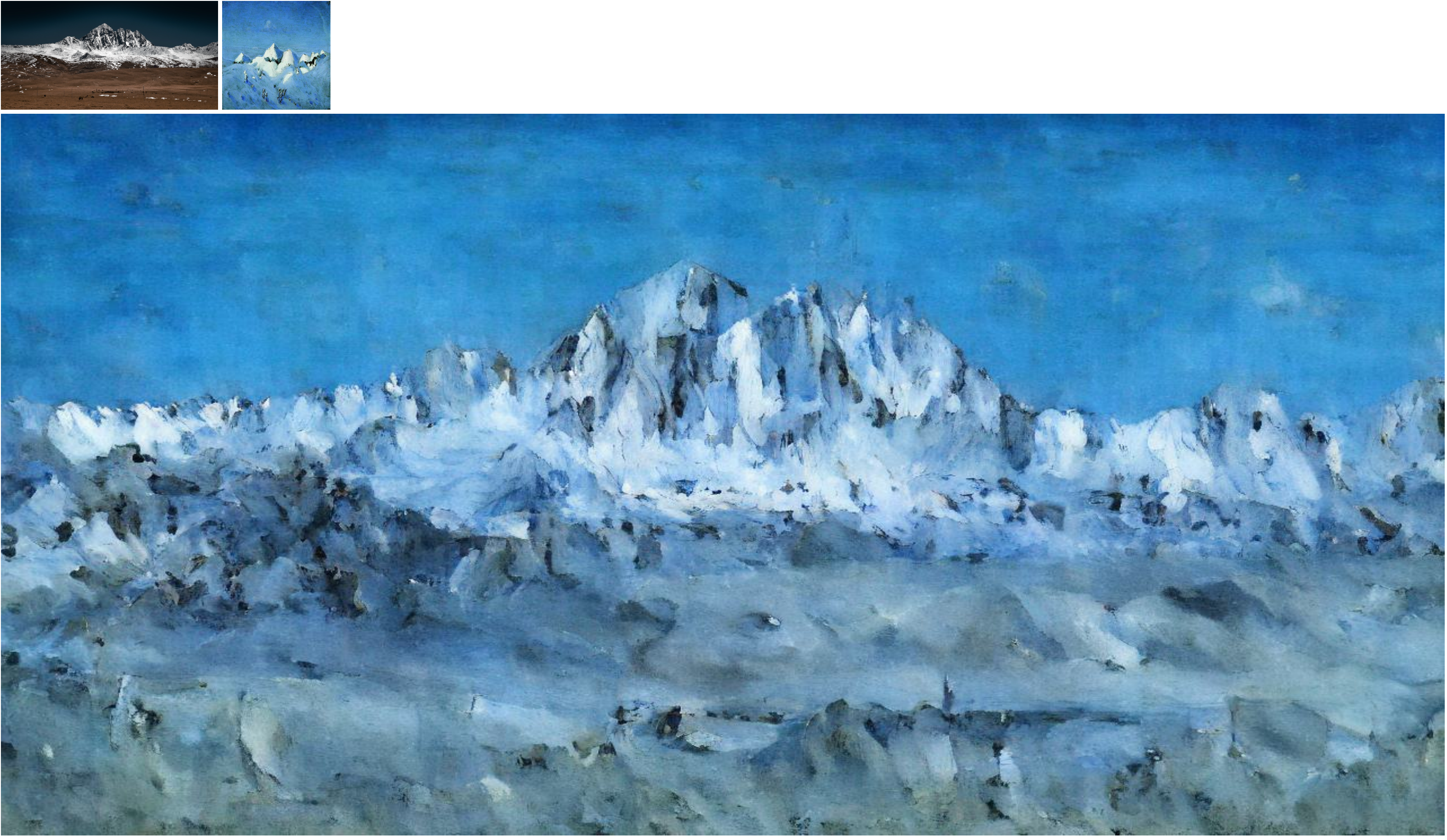}
\end{center}
   \caption{Samples with size $1280 \times 640$. 2D-CFG scales $s_{cnt}/s_{sty}$ from top to bottom: $0.25/1.5$ and $0.25/2.0$.}
\label{fig:high_resolution_2}
\end{figure}

\section{Details of Denoising Diffusion Probabilistic Models}
\label{sec:details_ddpm}
Given the data $x_0$, the Gaussian diffusion process, denoted by $q$, incrementally adds noise to $x_0$ to create noisy data at each timestep $t=1, ..., T$ as per the following equation:

\begin{equation}
    q(x_t|x_{t-1}) := \mathcal{N}(x_t; \sqrt{1 - \beta_t}, \beta_t\mathbf{I})
\end{equation}

\noindent
In this equation, ${\beta_t}^T_{t=1}$ represents the hyper variance schedule that dictates the extent of noise introduced at each timestep. We denote $\alpha_t := 1 - \beta_t$ and $\bar{\alpha}_t := \prod^t_{s=1}\alpha_s$ to express $q$ in an alternate form:

\begin{equation}
\begin{aligned}
    q(x_t|x_0) &= \mathcal{N}(x_t; \sqrt{\bar{\alpha}_t}x_0, (1 - \bar{\alpha}_t)\mathbf{I})\\
    &= \sqrt{\bar{\alpha}}_t x_0 + \epsilon \sqrt{1 - \bar{\alpha}_t}, \epsilon \sim \mathcal{N}(0, \mathbf{I})
    \label{eq:q_t_0}
\end{aligned}
\end{equation}

\noindent
This assists in an efficient sampling of $x_t$. With an appropriate variance schedule $\beta_t$ and sufficiently large $T$, the distribution $q(x_T)$ will converge to $\mathcal{N}(0, \mathbf{I})$. In such a case, given $x_T \sim \mathcal{N}(0, \mathbf{I})$, the Gaussian diffusion model $p_\theta$ seeks to approximate and parametrize the reverse distribution $q(x_{t-1}|x_t)$.
According to Sohl-Dickstein \etal \cite{DBLP:journals/corr/Sohl-DicksteinW15}, $q(x_{t-1}|x_t)$ can be treated as a diagonal Gaussian distribution as $T$ approaches infinity and $\beta_t$ tends to zero. Therefore, we can represent the parametrized $p_\theta$ as:

\begin{equation}
    p_\theta(x_{t-1}|x_t) := \mathcal{N}(x_{t-1}; \mu_\theta(x_t, t), \Sigma_\theta(x_t, t))
\end{equation}

\noindent
Here, $\mu_\theta(x_t, t)$ and $\Sigma_\theta(x_t, t)$ are learned deviation and mean. Ho \etal \cite{ddpm} observe that instead of directly optimizing the variational lower-bound for $p_\theta$ by learning both $\mu_\theta$ and $\Sigma_\theta$, the model can fix $\Sigma_\theta(x_t, t)$ to either $\beta_t\mathbf{I}$ or $\tilde{\beta}_t\mathbf{I}$, where $\tilde{\beta}_t := \frac{1 - \bar{\alpha}_{t-1}}{1 - \bar{\alpha}_t}\beta_t$ represents a rescaling of $\beta_t$.
Moreover, we can represent $\mu_\theta$ as:

\begin{equation}
    \mu_\theta(x_t, t) = \frac{1}{\sqrt{\alpha_t}}(x_t - \frac{1 - \alpha_t}{\sqrt{1 - \bar{\alpha}_t}}\epsilon_\theta(x_t, t))
\end{equation}

\noindent
With the prediction $\epsilon_\theta$ of the involved noise $\epsilon$ in Eq. \ref{eq:q_t_0}. The optimization goal can be transferred to minimize the difference between $\epsilon_\theta$ and $\epsilon$. This simplified objective is:

\begin{equation}
    \mathcal{L}_{simple} := \mathbb{E}_{x_0 \sim q(x_0), t \sim \mathcal{U}(\{1, ..., T\}), \epsilon \sim \mathcal{N}(0, \mathbf{I})}\left[\lVert \epsilon - \epsilon_\theta(x_t, t) \rVert_2^2\right]
\end{equation}

\noindent
The main optimization goal, therefore, is to align the predicted and actual noise terms as closely as possible.

\newpage
\section{Implementation Details}
We employed the MS-COCO dataset \cite{ms_coco} to train the partial conditional $\epsilon_\theta(z_t, z_c, \O_s)$, the model that exclusively conditions on content.
Meanwhile, the WikiArt dataset \cite{wikiart} was selected for training both $\epsilon_\theta(z_t, z_c, f_s)$ and $\epsilon_\theta(z_t, \O_c, f_s)$ due to its diverse artistic styles.
All the images used in the training process were randomly cropped into a 256x256 size.
Our model was trained on a single NVIDIA GeForce RTX 3080 Ti GPU.
Throughout the training process, we maintained an exponential moving average (EMA) of ArtFusion with a decay rate of 0.9999.
Unless otherwise specified, our results were sampled using the EMA model with 250 DDIM \cite{ddim} steps and setting the 2D-CFG scales as $s_{cnt}/s_{sty} = 0.6/3$.
The hyperparameters used for the architecture and training process of ArtFusion are detailed in Tab. \ref{tab:hyper_model} and \ref{tab:hyper_train}, respectively.
We did not conduct hyperparameter sweeps in this study.

\begin{table}[h]
\begin{center}
    \begin{tabular}{ccccccc}
        \toprule
        $z$-shape &                 $z_r$-shape         &     Channels  & Channel Multi. & Head Dim. & Embed. Dim. & Attn. Resolution \\
        \midrule
        $16 \times 16 \times 16 $ & $12 \times 16 \times 16 $ & 384    &   1, 2          & 64        & 1024        & 16, 8   \\
        \bottomrule
    \end{tabular}
\end{center}
\caption{Hyperparameters for the architecture of ArtFusion.}
\label{tab:hyper_model}
\end{table}

\begin{table}[h]
\begin{center}
    \begin{tabular}{cccccc}
        \toprule
        Diffusion Steps & Noise Schedule & Batch Size & Iterations & Optimizer & Learning Rate \\
        \midrule
        1000 &             Linear &        128 &          175K       &   AdamW &  $1e-4$  \\
        \bottomrule
    \end{tabular}
\end{center}
\caption{Hyperparameters for the training process of ArtFusion.}
\label{tab:hyper_train}
\end{table}

\newpage
\section{Inference Analysing}

\subsection{Sampling Steps}
\begin{figure}[h]
\begin{center}
   \includegraphics[width=.8\linewidth]{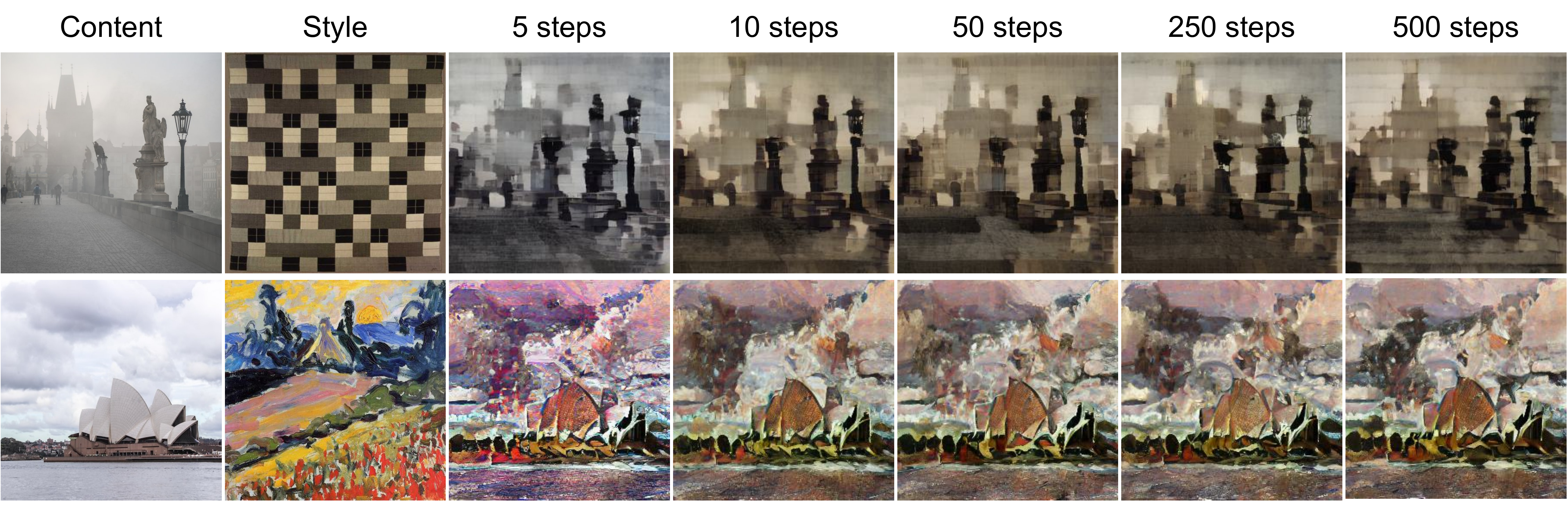}
\end{center}
   \caption{Impact of DDIM sampling steps on stylization outcomes. 10 sampling steps are enough for high-fidelity results.}
\label{fig:ddim}
\end{figure}

Figure \ref{fig:ddim} showcases a series of stylized results from varying DDIM \cite{ddim} steps.
Interestingly, our observations indicate that beyond 10 sampling steps, any additional steps have only a marginal improvement in the visual quality of the results.
This implies that despite our default setting of 250 steps, a reduction to merely 10 sampling steps does not lead to a noticeable deterioration in the quality of the output.

\subsection{Inference Time}
We evaluated the inference time of our model in comparison to other SOTA methods, as outlined in Table \ref{tab:time}. These comparisons utilized images of 256x256 resolution on an RTX 3080 Ti GPU.
DiffuseIT \cite{kwon2023diffusionbased}, another diffusion-based method, requires notably extended inference times compared to our model.
This increased time is due to DiffuseIT's reliance on DINO ViT \cite{dino}, which requires execution in both forward and backward directions during each sampling step to provide guidance.
On the other hand, our model maintains a competitive inference time, approximately $3\times$ as long as ArtFlow \cite{artflow} when sampling with 10 steps.
Considering the continued advancements in accelerated denoising inference processes, we expect the current efficiency gap to diminish in the near future.

\begin{table}[h]
\begin{center}
    \begin{tabular}{ccccccccccc}
        \toprule
            & Ours & Ours 10 steps & DiffuseIT & StyTr$^2$  & Styleformer & CAST     & IEST   & AdaAttn & ArtFlow & AdaIN\\
        \midrule
        time & 3.658s      &  0.196s    &    60.933s       &   0.051s     &  0.022s    & 0.018s  & 0.018s &  0.024s  &  0.063s & 0.017s \\
        \bottomrule
    \end{tabular}
\end{center}
\caption{Inference time comparison among SOTA methods.}
\label{tab:time}
\end{table}

\newpage
\section{Limitation}
\begin{figure}[h]
\begin{center}
   \includegraphics[width=.9\linewidth]{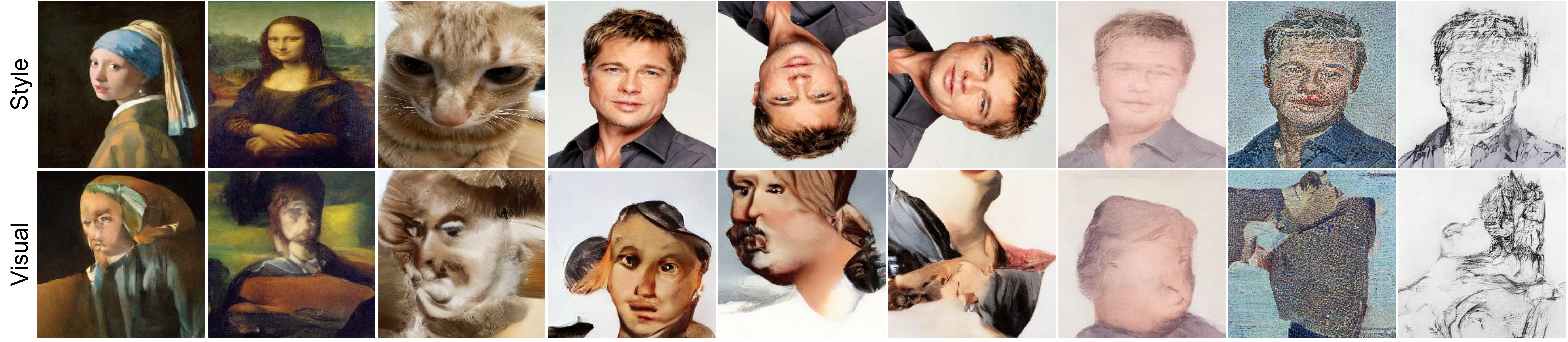}
\end{center}
   \caption{Visualization of style composition with human faces. It highlights cases of overfitting in style references that include face-like objects, without strong style patterns.}
\label{fig:face}
\end{figure}

Our model tends to overfit on the most recurring patterns in the WikiArt dataset, namely, frontal human faces. Among various art categories, portraits represent $15\%$ of the entire dataset.
This overfitting is evident in multiple cases, as showcased in style visualizations from $\epsilon_\theta(z_t, \O_c, f_s)$ in Fig. \ref{fig:face}.
For instance, images comprising human faces or objects with human-like attributes, as well as images featuring upside-down faces, appear to predominantly overfit on horizontal facial patterns (as evident in the $1^{st} - 6^{th}$ columns).
However, this issue seems to alleviate or even vanish when images incorporate other discernible style features (as observable when comparing the $4^{th}$ and $7^{th} - 9^{th}$ columns).
Moving forward, we intend to overcome this limitation through the implementation of more robust data augmentation strategies.

\section{Additional Qualitative Results}

\noindent\textbf{High Resolution}.
We assess the scalability of ArtFusion by testing it on higher-resolution content images, while keeping the size of the style images at $256 \times 256$ (refer to Fig. \ref{fig:high_resolution} and \ref{fig:high_resolution_2}).
ArtFusion adeptly scales to high resolutions without any fine-tuning, preserving high levels of detail and yielding aesthetically pleasing results.

\begin{figure}[H]
\begin{center}
   \includegraphics[width=.7\linewidth]{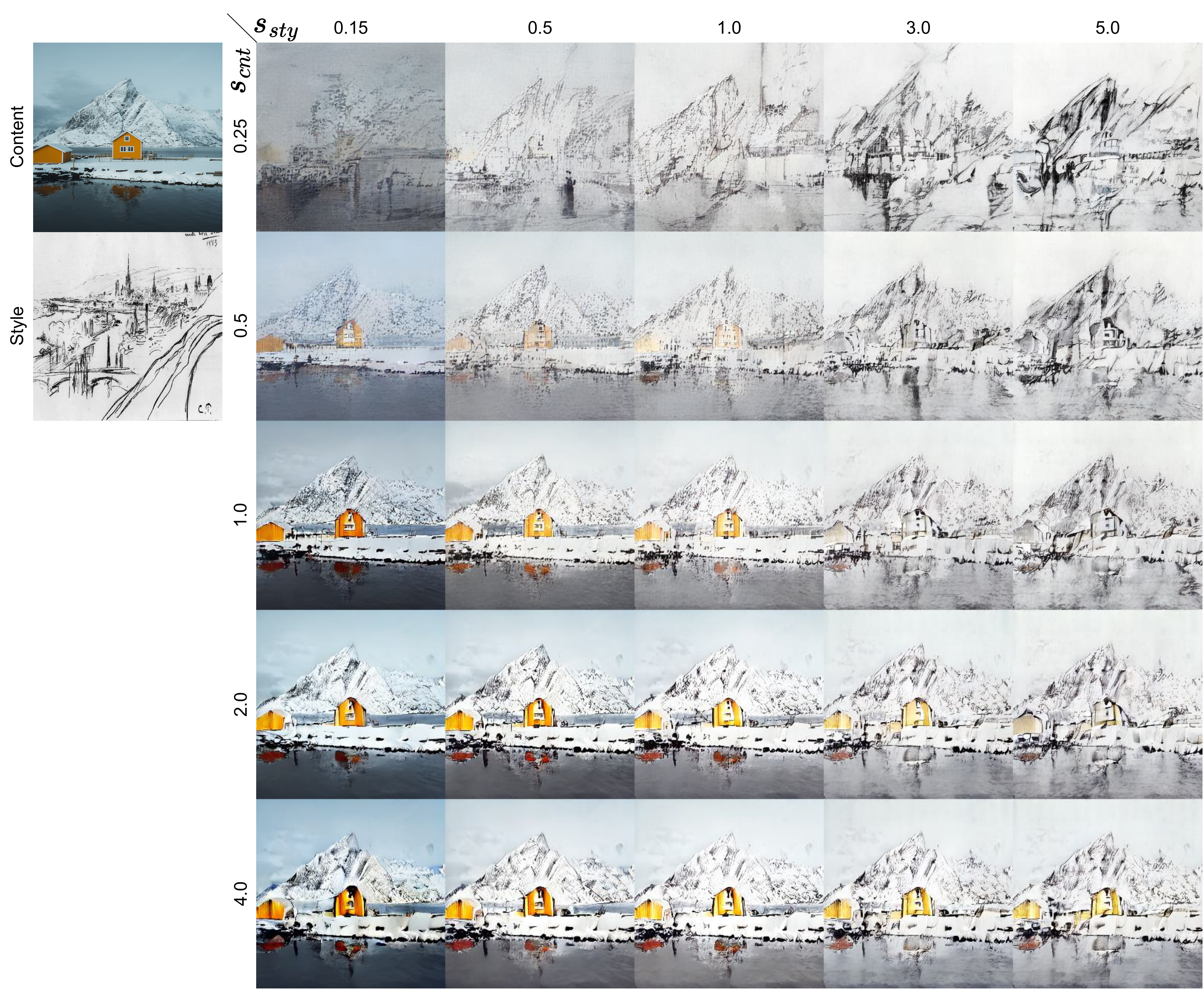}

    \vspace{0.5cm}
   
   \includegraphics[width=.7\linewidth]{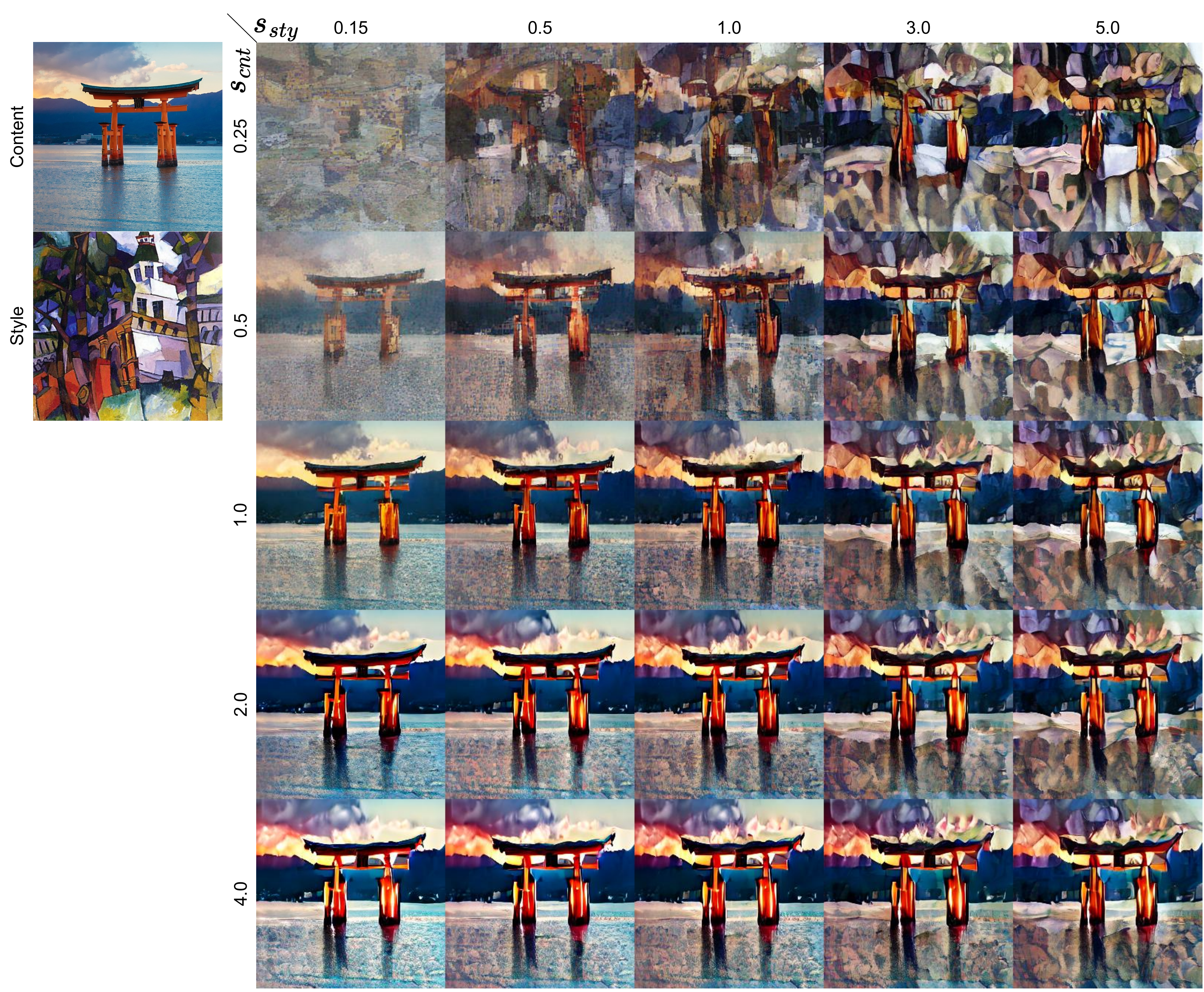}
\end{center}
   \caption{Additional two-dimensional classifier-free guidance results.}
\label{fig:two_dim_more}
\end{figure}

\begin{figure}[H]
\begin{center}
   \includegraphics[width=1.\linewidth]{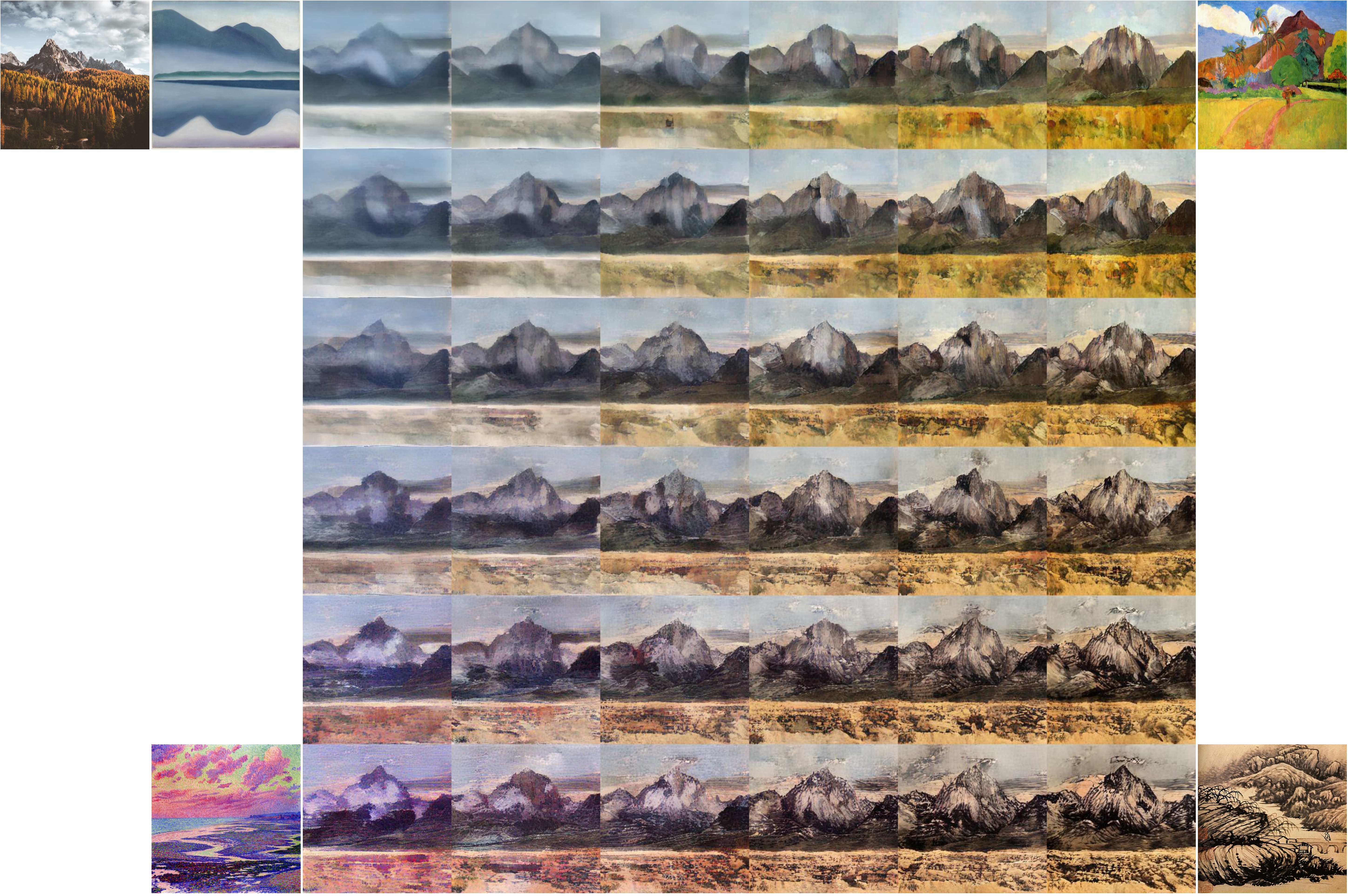}
\end{center}
   \caption{Interpolation results between four styles.}
\label{fig:interpolate_four}
\end{figure}

\begin{figure}[H]
\begin{center}
   \includegraphics[width=1.\linewidth]{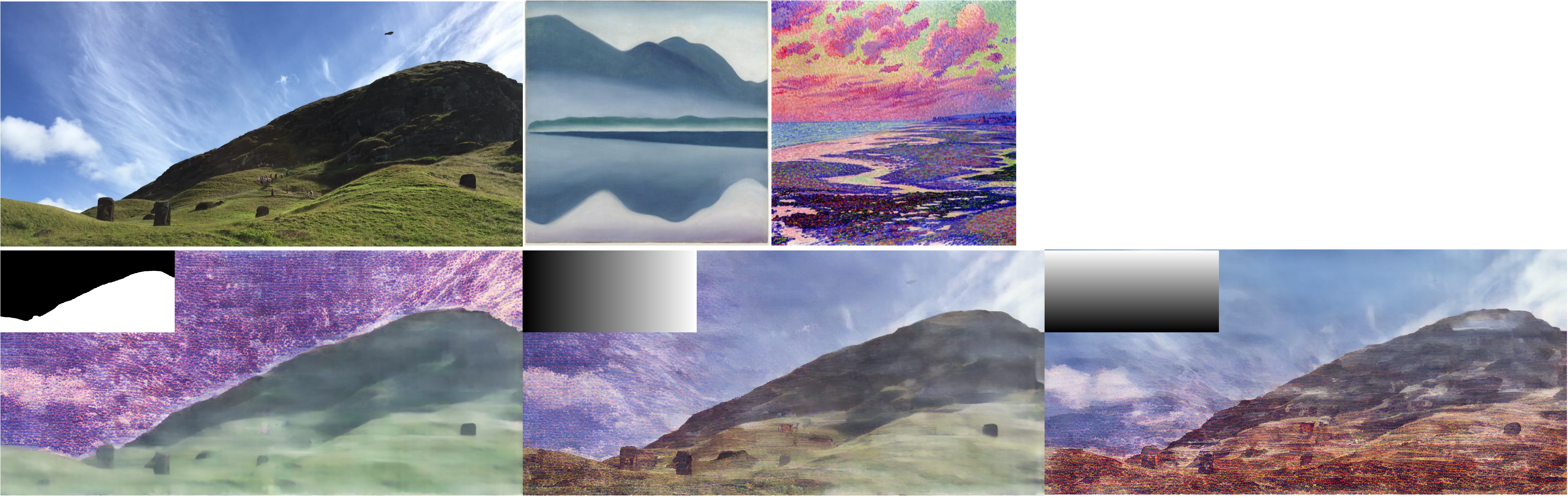}
\end{center}
   \caption{Results of spatial control with size $1024 \times 480$. The first row is the content and style images. The second row showcases the spatial control results along with the corresponding gradient masks.}
\label{fig:spatial}
\end{figure}

\noindent\textbf{Manipulation}.
The controlling capabilities of our model transcend conventional limits, as demonstrated by the 2D-CFG samples (refer to Fig. \ref{fig:two_dim_more}) and style interpolations between four styles (see Fig. \ref{fig:interpolate_four}).
Additionally, we demonstrate how the application of gradient masks in style interpolation allows for precise spatial control over style proportions in Fig. \ref{fig:spatial}.
The suite of manipulation methods we have introduced enhances the versatility and practicality of AST for real-world applications.

\begin{figure}[H]
\begin{center}
   \includegraphics[width=1.\linewidth]{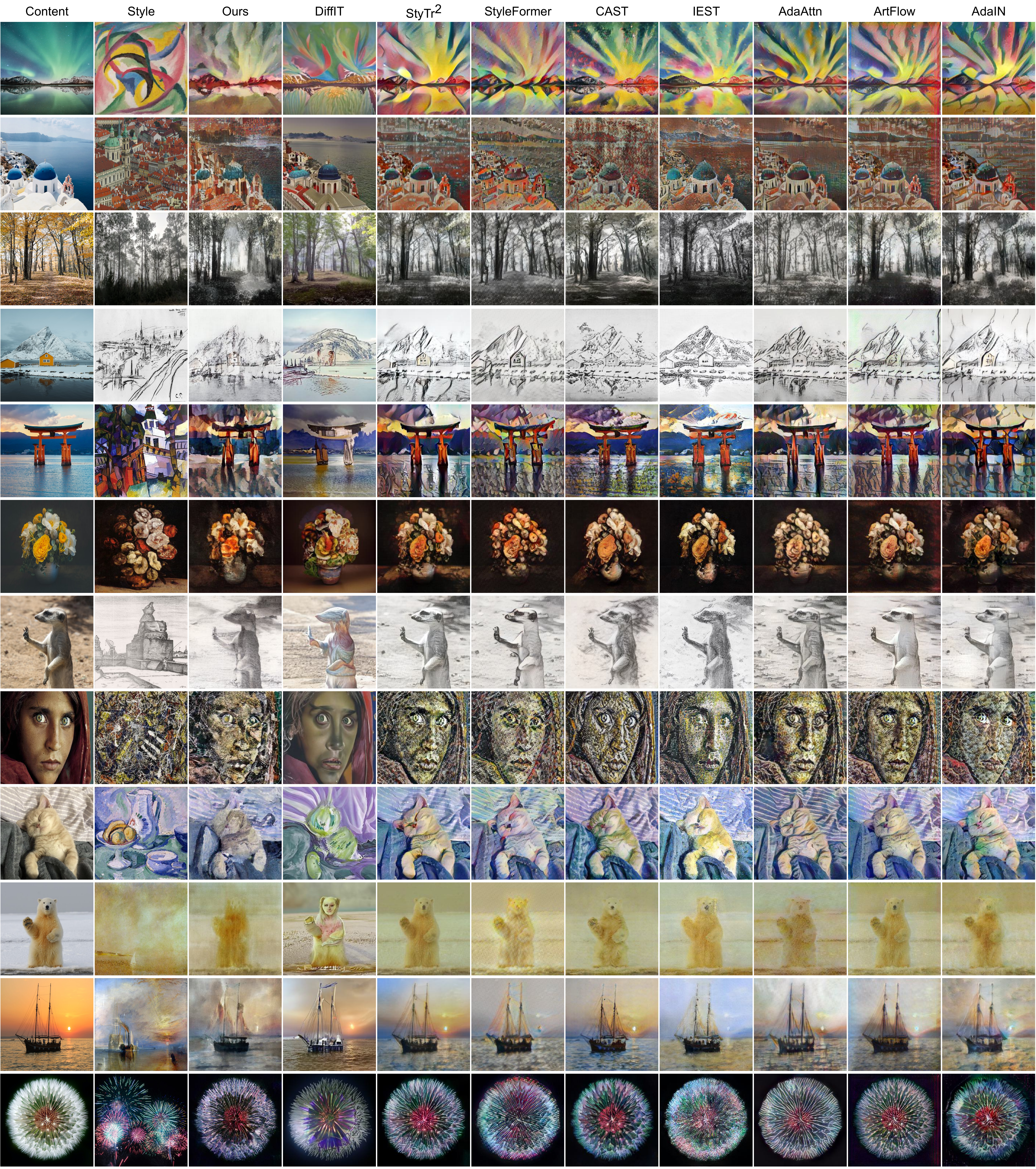}
\end{center}
   \caption{Additional comparison with SOTA results.}
\label{fig:compares_more}
\end{figure}

\noindent\textbf{Comparison}.
Additional comparison with SOTA approaches \cite{kwon2023diffusionbased, deng2021stytr2, styleformer, cast, iest, adaattn, artflow, adain} is illustrated in Fig. \ref{fig:compares_more}. We encourage a closer examination of these figures, as the zooming-in details truly showcase the superior performance of our model. It is here where the real strengths of ArtFusion shine – in its fine-grained details.

\end{document}